\documentclass{article} % For LaTeX2e
\usepackage{arxiv}

\newtoggle{arxiv}
\toggletrue{arxiv}

\usepackage{amsmath,amsfonts,bm}

\def\eqref#1{equation~\ref{#1}}
\def\1{\bm{1}}

\DeclareMathAlphabet{\mathsfit}{\encodingdefault}{\sfdefault}{m}{sl}
\SetMathAlphabet{\mathsfit}{bold}{\encodingdefault}{\sfdefault}{bx}{n}

\usepackage{microtype}
\usepackage{graphicx}
\usepackage{subfigure}
\usepackage{float}
\usepackage{booktabs} % for professional tables
\usepackage{hyperref}
\usepackage{bbm}
\usepackage{tabularx}
\usepackage{multirow}
\usepackage{siunitx}
\usepackage{makecell}
\usepackage{listings}
\usepackage{placeins}
\usepackage{graphicx}
\usepackage{wrapfig}
\usepackage{caption}   % Required for \captionsetup

\usepackage{adjustbox} % For table resizing if needed

\usepackage{enumitem}

\usepackage{amsmath}
\usepackage{amssymb}
\usepackage{mathtools}
\usepackage{amsthm}
\usepackage{pifont}
\newcommand{\xmark}{\ding{55}}%

\newcommand{\fast}[1]{$\text{fast}_{#1}$}

\lstdefinelanguage{CUDACPP}{
  language=C++,
  morekeywords={__global__, __host__, __device__, __shared__, blockIdx, blockDim, threadIdx, gridDim},
  morecomment=[l][\color{magenta}]{\#},
}
\theoremstyle{plain}
\usepackage[textsize=tiny]{todonotes}

\iftoggle{arxiv}{
    \usepackage[numbers,sort]{natbib}
    \setlength{\textwidth}{6.5in}
    \setlength{\textheight}{9in}
    \setlength{\oddsidemargin}{0in}
    \setlength{\evensidemargin}{0in}
    \setlength{\topmargin}{-0.5in}
    \newlength{\defbaselineskip}
    \setlength{\defbaselineskip}{\baselineskip}
    \setlength{\marginparwidth}{0.8in}
}

\newcolumntype{Y}{>{\hsize=.7\hsize}X}
\newcolumntype{Z}{>{\hsize=1.3\hsize}X}

\makeatletter
\def\@copyrightspace{\relax}
\makeatother

\makeatletter
\def\@myauthornotes{}
\def\myauthornote#1{%
  \if@ACM@anonymous\else
    \g@addto@macro\addresses{}%
    \g@addto@macro\@myauthornotes{%
      \stepcounter{footnote}\footnotetext{#1}}%
  \fi}
\makeatother

\iftoggle{arxiv}{
    \title{
        KernelBench: Can LLMs Write Efficient GPU Kernels?
        \footnotetext{*Equal Contribution. Correspondence: \texttt{aco@stanford.edu, simonguo@stanford.edu}}
    }
    \usepackage{authblk}
    \author[1,*]{Anne Ouyang}
    \author[1,*]{Simon Guo}
    \author[1]{Simran Arora}
    \author[2]{Alex L. Zhang}
    \author[1]{William Hu}
    \author[1]{Christopher Ré}
    \author[1]{Azalia Mirhoseini}
    
    \affil[1]{Stanford University}
    \affil[2]{Princeton University}\vspace{4pt}
}

\begin{document}

\maketitle


% \begin{icmlauthorlist}
% \icmlauthor{Anne Ouyang}{equal,stanford}
% \icmlauthor{Simon Guo}{equal,stanford}
% \icmlauthor{Simran Arora}{stanford}
% \icmlauthor{Alex Zhang}{princeton}
% \icmlauthor{William Hu}{stanford}
% \icmlauthor{Christopher Ré}{stanford}
% \icmlauthor{Azalia Mirhoseini}{stanford}
% \end{icmlauthorlist}

% \icmlaffiliation{stanford}{Department of Computer Science, Stanford University, Stanford, California, USA}
% \icmlaffiliation{princeton}{Department of Computer Science, Princeton University, Princeton, New Jersey, USA}

% \icmlcorrespondingauthor{Anne Ouyang}{aco@stanford.edu}
% \icmlcorrespondingauthor{Simon Guo}{simonguo@stanford.edu}
% \vskip 0.3in


\begin{abstract}
% Keep your abstract brief and self-contained, one paragraph and roughly 4--6 sentences.
Efficient GPU kernels are crucial for building performant machine learning architectures, but writing them is a time-consuming challenge that requires significant expertise; therefore, we explore using language models (LMs) to automate kernel generation. We introduce \textbf{KernelBench}, an open-source framework for evaluating LMs' ability to write fast and correct kernels on a suite of 250 carefully selected PyTorch ML workloads. KernelBench represents a real-world engineering environment and making progress on the introduced benchmark directly translates to faster practical kernels. We introduce a new evaluation metric \fast{p}, which measures the percentage of generated kernels that are functionally correct and offer a speedup greater than an adjustable threshold $p$ over baseline. Our experiments across various state-of-the-art models and test-time methods show that frontier reasoning models perform the best out of the box but still fall short overall, matching the PyTorch baseline in less than 20\% of the cases. While we show that results can improve by leveraging execution and profiling feedback during iterative refinement, KernelBench remains a challenging benchmark, with its difficulty increasing as we raise speedup threshold $p$.

%the previous version
%Efficient GPU kernels are crucial for building performant machine learning architectures, but writing them is a time-consuming challenge that requires significant expertise; therefore, we explore using language models (LMs) to automate kernel generation. We introduce \textbf{KernelBench}, a novel framework with a benchmark consisting of 250 diverse PyTorch ML workloads representative of a real-world kernel engineering environment. KernelBench targets the specialized domain of kernel generation requiring code to be both correct and fast, and making progress on this benchmark directly translates to faster kernels applicable in the real world. We introduce a new evaluation metric \fast{p}, which measures the percentage of generated kernels that are functionally correct and offer a speedup greater than an adjustable threshold $p$ over baseline. Our experiments across various state-of-the-art models and test-time methods show that frontier reasoning models perform the best out of the box but still fall short overall, matching the PyTorch baseline in less than 20\% of the cases. While we show that results can improve by leveraging execution and profiling feedback in multi-turn interactions, KernelBench remains a challenging benchmark as there is still a large headroom for improvement, especially as we increase our target threshold $p$.  

%however, there remains room for improvement, especially as we increase the threshold $p$ in our evaluation metric. 

% We propose fully automated programmatic evaluation methods
\end{abstract}

\label{submission}

\section{Introduction}
AI relies on efficient GPU kernels to achieve high performance and cost and energy savings; however, developing kernels remains challenging. 
% FlashAttention3~\cite{dao2024flashattention3}. 
There has been a Cambrian explosion of ML architectures~\cite{tay2022efficient, peng2023rwkv, dao2024transformers}, but their available implementations routinely underperform their peak potential. We are seeing a proliferation of AI hardware ~\cite{nvidia2017nvidia, nvidia2020nvidia, nvidia2022nvidia, jouppi2023tpuv4opticallyreconfigurable, groq-chip, cerebras-wse, graphcore-ipu}, each with different specs and instruction sets, and porting algorithms across platforms is a pain point. A key example is the FlashAttention kernel \cite{dao2022flashattention}, which is 
% for Transformers 
% a memory efficient implementation of the attention mechanism 
crucial for running modern Transformer models –– the initial kernel released in 2022, five years after the Transformer was proposed; it took two more years from the release of NVIDIA Hopper GPUs to transfer the algorithm to the new hardware platform. We explore the question: \textit{Can language models help write correct and optimized kernels?}

\begin{figure*}[!ht]
    \centering
    \includegraphics[width=\textwidth]{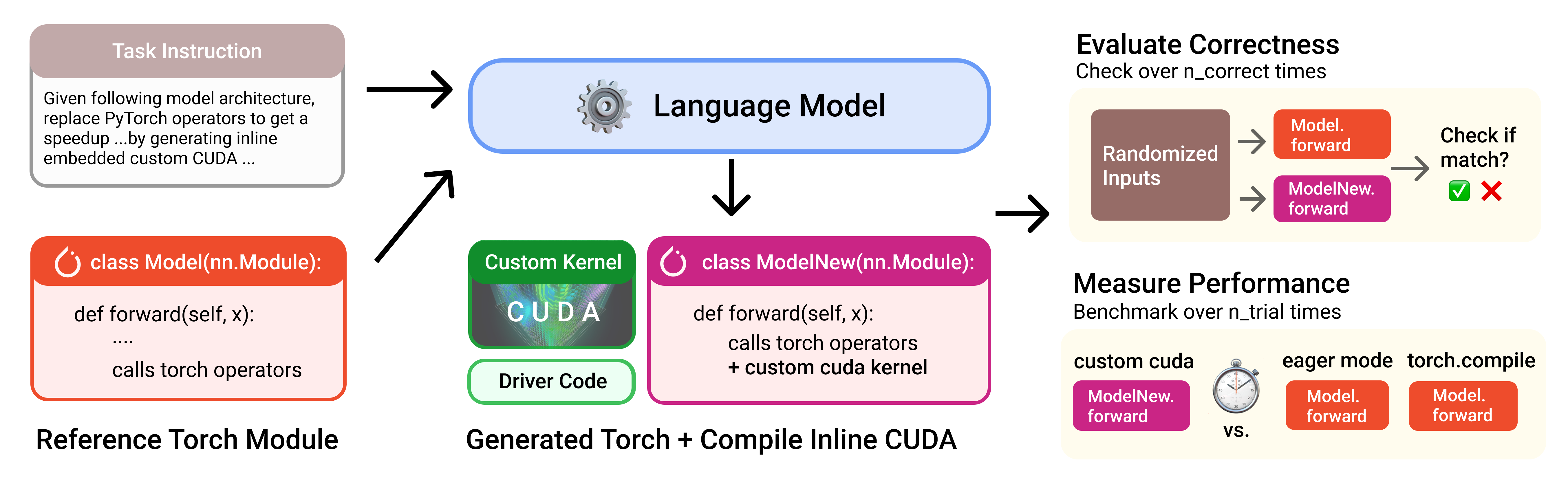} % Use \textwidth to span both columns
    \vspace{-20pt}
    \caption{\textbf{KernelBench evaluates LMs' ability to generate performant GPU Kernels}. Overview of tasks in KernelBench: KernelBench tasks LMs with generating optimized CUDA kernels for a given target PyTorch model architecture and conducts automated evaluation}
    \vspace{-10pt}
    \label{fig:kernelbench-workflow}
\end{figure*}

AI engineers use a rich set of information when developing kernels and it is not clear whether language models (LMs) can mimic the workflow. They use compiler feedback, profiling metrics, hardware-specific specs and instruction sets, and knowledge of hardware-efficiency techniques (e.g., tiling, fusion). They can use programming tools ranging from assembly (e.g., PTX as in ~\citet{deepseekv3}) to higher-level libraries (ThunderKittens~\cite{spector2023thunderkittens}, Triton~\cite{triton}). Compared to existing LM code generation workloads~\cite{yang2024swe}, kernel writing requires a massive \textit{amount} and \textit{diversity} of information. We first design an environment that reflects the typical AI engineer's workflow and supports providing LMs with this rich information. The environment should:
% that not only enables LMs to optimize kernel generation efficiently but also ensures that the evaluation is meaningful and reflects LMs' full potential.
% This 
% raises the 
% requires careful navigation of a large design space with the 
% following key considerations:

\begin{itemize}[itemsep=0.1pt,topsep=0pt,leftmargin=*]
    \item \textbf{Automate} the AI engineer's workflow. The model should have full flexibility to decide which operators to optimize and how to optimize them.
    \item Support a \textbf{diverse} set of AI algorithms, programming languages, and hardware platforms.
    \item Make it \textbf{easy to evaluate} both performance and functional correctness of LM generations, ideally in a programmatic way. It should also capture profiling and execution information from generated kernels. 
    % process is desirable, as it enables reproducible large-scale benchmarking. 
\end{itemize}

\noindent We introduce \textbf{KernelBench} to generate and evaluate kernels, which addresses the above considerations. KernelBench tests LM optimizations on three levels of AI workloads:
\begin{enumerate}[itemsep=0.1pt,topsep=0pt,leftmargin=*]
    \item \textbf{Individual operations:} We include various AI operators, including matrix multiplies, convolutions, activations, norms, and losses. While PyTorch already uses expert-optimized closed-source kernels, making this a potentially challenging baseline, it is valuable if LMs can generate open-source kernels for the operations.
    \item \textbf{Sequence of operations:} We provide problems that contain 3-6 individual operations together (e.g. a mainloop operator like matmul followed by pointwise operators like ReLU and Bias). This enables evaluating the models' ability to fuse multiple operators.
    % , an important optimization for kernels.
    \item \textbf{End-to-end architectures:} We select architectures from popular AI repositories on Github including \texttt{pytorch}, \texttt{huggingface/transformers}, and \texttt{huggingface/pytorch-image-models}. These architectures contain many operations.
\end{enumerate}

\noindent Mimicking an AI researcher's workflow, the LM takes PyTorch reference code as input and outputs an optimized version of the code.
% while maintaining functional correctness. 
Similar to the human kernel development process, our environment enables the LM to iterate with compiler and profiler feedback to refine performance. The LM is free to use any programming language and decide both \textit{which parts} of the PyTorch code to optimize, and \textit{how} to optimize them. Our pipeline allows us to feed diverse information to the LMs, including hardware-specific information, example kernels, and compiler/profiler feedback.
% The design is flexible in accommodating diverse programming languages and libraries because the LM can implement any approach as long as it integrates with PyTorch, which is a typical workflow in AI research.

% Our set-up allows us to programmatically measure success by comparing model output to the reference for correctness and to the PyTorch baseline for speed. We unify these metrics by proposing the $\mathbf{fast_p}$ metric: the percentage of correct kernels that outperform the PyTorch reference by $ > \mathbf{p}$ speedup, where speedup the ratio of PyTorch baseline wallclock time to generated kernel time. An advantage of this metric is that its adaptability  –– increasing the threshold $p$ allows us to progressively increase the evaluation's difficulty.

We observe that frontier and open-source models perform poorly out-of-the-box on KernelBench, with OpenAI-o1 and DeepSeek-R1 matching the PyTorch Eager baseline on $<20\%$ of the tasks. These model-generated kernels greatly suffer from execution errors, functional correctness issues, and are unable to perform platform-specific optimizations. To identify areas for improvement, we conduct a series of experiments and analysis, and find that:

% \begin{itemize}[itemsep=0.1pt,topsep=0pt,leftmargin=*]
% \begin{enumerate}[leftmargin=*, itemsep=0pt] % or 
\begin{enumerate}[itemsep=0.1pt,topsep=0pt,leftmargin=*]
    \item \textit{\textit{Writing functionally }\textit{\textbf{correct}}\textit{ kernels remains challenging for models:} }while models are able to fix execution failures through either reasoning or multiple attempts, they struggle to produce functionally correct code. Furthermore, we observe a trade-off between LMs attempting more complex optimizations / niche hardware instructions (e.g., tensor core \texttt{wmma}) and producing error-free kernels. We hypothesize this is due to CUDA being a low-resource language in open-source training data, only $0.073\%$ of popular code corpus The Stack v1.2 \cite{li2023starcodersourceyou, kocetkov2022stack3tbpermissively}.
    
    \item \textit{\textit{Models demonstrate potential to produce }\textit{\textbf{performant}}\textit{ kernels via optimizations:}} We observe a few instances where LMs make algorithmic improvements -- e.g., exploiting sparsity, operator fusion, and utilizing hardware features. We notice more of such instances when we explicitly condition the LM on hardware information (e.g., bandwidth and TFLOP specs) and demonstrations of hardware optimization techniques (e.g., tiling, fusion). While these capabilities remain nascent, LMs do \textbf{demonstrate potential} for generating performant kernels. 
    \item \textit{Leveraging}\textit{\textbf{ feedback}}\textit{ is important for reducing execution errors and discovering faster solutions:} By providing execution results and profiler feedback to the LM in context, the kernel quality significantly improves after multiple refinements from $12\%$, $36\%$, and $12\%$ in \fast{1} to $43\%$, $72\%$, and $18\%$ respectively.
\end{enumerate}

\noindent Our findings highlight the technical challenges we need to solve in order to adopt LMs for kernel writing. These include but are not limited to: how to improve LM performance in a low-resource data regime, and how to select from the rich set of information we can provide to models. To address these challenges, we contribute (1) \textbf{an open-source framework} to study LM kernel generation with a comprehensive suite of evaluation problems and (2) \textbf{analysis of where current LMs stand} and how to realize a future of efficient kernels generated by models. 

\vspace{-2mm}
\section{Related Works}
\textbf{Kernel libraries and compilers.} We evaluate existing approaches for kernel programming along the dimensions of automation, breadth, and performance. Mainstream kernel programming libraries like cuDNN ~\cite{nvidiacudnn}, CUTLASS~\cite{nvidia2017cutlass}, and Apple MLX~\cite{applemlx} are hardware-specific and demand substantial engineering effort from human experts. Other libraries, like ThunderKittens~\cite{spector2023thunderkittens} and  Triton~\cite{triton}, successfully help AI researchers write a breadth of fast and correct kernels~\cite{arora2024simple, yang2024fla}, but still require human programming effort. Compiler-based tools, like torch.compile~\cite{paszke2019pytorchimperativestylehighperformance} and FlexAttention~\cite{he2024flexattention}, automatically provide a narrow slice of optimizations. In contrast to these efforts, we ask if LMs can automatically generate performant kernels for a breadth of AI workloads. 
\\\\
\noindent \textbf{LLMs for performance-optimized code generation.} In the past year, there have been several efforts to build LMs that can automate algorithmic coding~\cite{chen2021evaluatinglargelanguagemodels, shi2024languagemodelssolveolympiad, Li_2022}, resolving GitHub issues~\cite{yang2024swe,yang2024swebenchmultimodalaisystems}, and domain-specific coding~\cite{yin2022naturallanguagecodegeneration,lai2022ds1000naturalreliablebenchmark}. While these works focus on producing correct and functional code, subsequent works have explored LMs' ability to produce solutions with better \textit{algorithmic and asymptotic efficiency} ~\cite{nichols2024performancealignedllmsgeneratingfast,waghjale-etal-2024-ecco}. KernelBench focuses on \textit{wall-clock efficiency}. LMs generate high-performance computing (HPC) code, which requires an understanding of the underlying hardware features and device instruction set, and common performance characteristics of parallel processors. 

Existing works in the space of HPC code generation have evaluated LM performance on translating arbitrary code samples from C++ to CUDA~\cite{tehranijamsaz2024coderosetta,pmlr-v162-wen22b} or generating well-known, low-level kernels such as GEMMs~\cite{valerolara2023comparingllama2gpt3llms, wijk2024rebenchevaluatingfrontierai}. KernelBench instead curates a set of 250 diverse kernels from real-world, modern deep learning workloads, many of which do not have existing human-written implementations — in other words, solving KernelBench tasks are immediately beneficial for real deep learning workloads. 

\vspace{-2mm}
\section{KernelBench: A Framework for AI Kernel Generation}
KernelBench is a new framework for evaluating the ability of language models to generate performant kernels for a breadth of AI workloads.  In this section, we describe the task format, contents, and evaluation metric. 

\subsection{KernelBench Task Format}
KernelBench contains 250 tasks representing a range of AI workloads, and is easily extensible to new workloads. The end-to-end specification for a task is illustrated in \autoref{fig:kernelbench-workflow} and described below. 
\\\\
\noindent \textbf{Task input:} Given an AI workload, the input to the task is a reference implementation written in PyTorch. Mimicking an AI researcher's workflow, the PyTorch code contains a class named \texttt{Model} derived from \texttt{torch.nn.Module()}, where the standard \texttt{\_\_init\_\_} and \texttt{forward()} functions (and any helper functions) are populated with the AI workload's PyTorch operations. 

AI algorithms generally operate on large tensors of data. The optimal kernel for a workload depends on the size and data type (e.g., BF16, FP8) of the tensor. Therefore, each task additionally contains functions \texttt{get\_inputs()} and \texttt{get\_init\_inputs()}, which specify the exact input tensors that the kernel needs to handle.
\\\\
\noindent \textbf{Task output:} Given the input, the LM needs to output a new class named \texttt{ModelNew} derived from \texttt{torch.nn.Module()}, which contains custom optimizations. For example, the LM can incorporate in-line kernel calls during the \texttt{forward()} function using the CUDA-C extension in PyTorch.

In order to succeed, the LM needs to identify (1) which operations in the \texttt{Model} class would benefit most from optimizations, and (2) how to optimize those operations. The LM can use any hardware-efficiency techniques such as fusion and tiling or specialized instructions (e.g., tensor cores) and any programming library (e.g., PTX, CUDA, CUTLASS, Triton, ThunderKittens).

\subsection{Task Selection}
The 250 tasks in KernelBench are partitioned into three levels, based on the number of primitive operations, or PyTorch library functions, they contain:
% A primitive operation is equivalent to a PyTorch library function:
\begin{itemize}[itemsep=0.1pt,topsep=0pt,leftmargin=*]
    \item \textbf{Level 1 (100 tasks): Single primitive operation.} This level includes the foundational building blocks of AI (e.g. convolutions, matrix-vector and matrix-matrix multiplications, losses, activations, and layer normalizations). 

    Since PyTorch makes calls to several well-optimized and often closed-source kernels under-the-hood, it can be challenging for LMs to outperform the baseline for these primitive operations. However, if an LM succeeds, the open-source kernels could be an impactful alternative to the closed-source (e.g., CuBLAS~\cite{nvidia2023cublas}) kernels.
    
    \item \textbf{Level 2 (100 tasks): Operator sequences.} This level includes AI workloads containing multiple primitive operations, which can be fused into a single kernel for  improved performance (e.g., a combination of a convolution, ReLU, and bias).

    Since compiler-based tools such as the PyTorch compiler are effective at fusion, it can be challenging for LMs to outperform them. However, LMs may propose more complex algorithms compared to compiler rules. 

    % To generate these problems, we programmatically combine random choices of mainloop operators (e.g., matmul, conv) with random choices of epilogue operators (e.g., activations, norms, reductions). 
    % A script randomly selects one mainloop operator and 2–5 epilogue operators to create a specification, which is paired with a one-shot example for LLMs to generate PyTorch code.  On average, each problem contains 4 operations.
    
    \item \textbf{Level 3 (50 tasks): Full ML architectures.} This level includes architectures that power popular AI models, such as AlexNet and MiniGPT, collected from  popular PyTorch repositories on GitHub.
    
    Given the scale of modern models, it is critical to use kernels when running training and inference. Unfortunately, it has been difficult for the AI community to generate performant kernels. For instance, it took 5 years from the release of the Transformer architecture~\cite{vaswani2018attention} to obtain performant kernels~\cite{dao2022flashattention}, let alone today's many new architectures. Peak performance kernels for these architectures require algorithmic modifications that are often beyond the scope of a compiler. 
    
    % Level 3 includes a mix of LLM-generated well-known ML architectures (e.g., AlexNet) and real-world architectures like MiniGPT collected from the popular pytorch repositories on GitHub. These architectures have been cleaned up by removing extraneous details, such as argument parsers, training loops, dataset preprocessing, and non-essential components, leaving only the inference forward pass and necessary helper functions.
\end{itemize}

\noindent We reiterate that each task contains a meaningful set of AI primitive operations or architectures, such that LM success on the task can directly lead to real world impact.   

% \textbf{Problem sources} The tasks in KernelBench are created through a mix of manual authoring, LLM or script generation, and collection from sources like GitHub, with all tasks manually cleaned and verified for quality. Common ML operators are manually curated and implemented for level 1. At Level 2, mainloop operators (e.g., matmul, conv) and epilogue operators (e.g., activations, norms, reductions) are combined programmatically. A script randomly selects one mainloop operator and 2–5 epilogue operators to create a specification, which is paired with a one-shot example for LLMs to generate PyTorch code. 

% We also provide 20 additional tasks in an aspirational level 4. Unlike the previous levels where the tasks are self-contained, level 4 tasks call the Huggingface library such that optimizations will require changing the source code. Level 4 is currently a far-reaching objective, and we do not provide evaluations in this paper; however, we believe that this level could inform advancing the capabilities of LLMs to optimize complex real-world ML codebases. 

\subsection{Metric Design} \label{sec-metric-design}

We describe the evaluation approach for KernelBench and how we compare the success of different LMs. 

\paragraph{Evaluation approach}
KernelBench is an evaluation-only benchmark. We do not provide ground truth kernels for the tasks since we imagine users benchmarking on a variety of hardware platforms (including new platforms), input types, and workloads. However, by design, KernelBench is \textit{automatically verifiable}. Given a task, we randomly generate input tensors of the prescribed shape and precision and collect the PyTorch \texttt{Model} output. We can evaluate whether LM generations are correct and fast as follows:

\begin{enumerate}[itemsep=0.1pt,topsep=0pt,leftmargin=*]
    \item \textbf{Correctness} 
    % We check whether the output of the forward-pass from \texttt{Model} matches \texttt{ModelNew}, given the tensors input shapes and data types defined in the task. 
    We compare the \texttt{Model} output to the LM-generated \texttt{ModelNew} output. 
    % This approach is sufficient because the benchmark focuses on specialized code for specific problems, rather than generalized solutions. Custom kernels are only required to pass the test cases for their respective problems. Since the deep learning kernels do not involve control flow or branch-specific execution paths, correctness can be determined by verifying the numerical outputs against the reference implementation. 
    We evaluate on 5 random inputs per problem (detailed in Appendix~\ref{appendix:eval-methods-and-baselines}).
    % shows that correctness is either consistently achieved for all cases or fails completely on randomized inputs. 
    % This allows for a simple yet reliable mechanism to assess the functional accuracy of the generated kernels.
    \item \textbf{Performance} We compare the wall-clock execution time of \texttt{Model} against \texttt{ModelNew} using repeated trials to account for timing variations. 
    % We compare
\end{enumerate}

\paragraph{Comparing LMs on KernelBench}
Some LMs may generate a small number of correct kernels that are very fast, while other LMs generate a large number of correct kernels that are quite slow. Here, we explain our proposed unified metric for ranking LM quality on KernelBench. 

To capture both axes of correctness and performance, we introduce a new metric called \fast{p}, which is defined as the fraction of tasks that are both correct and have a speedup (computed as the ratio of PyTorch wall-clock time to generated kernel time) greater than threshold $p$. Formally:
\begin{align*}
\text{fast}_p = \frac{1}{N} \sum_{i=1}^N \mathbbm{1}(\text{correct}_i \land \left\{ \text{speedup}_i > p \right \}),
\vspace{-.1in}
\end{align*} 
where \fast{0} is equivalent to the LM's correctness rate, as it measures the fraction of tasks for which the LM code is functionally correct regardless of its speed.

By adjusting the threshold parameter $p$, we enable evaluation of kernel performance at different speedup thresholds and capture the speedup distributions. For our evaluations, we focus on $p=1$ as a starting point, with the possibility of increasing $p$ as future methods for kernel generation improve. Additionally, using $p<1$ for training is valuable, since PyTorch relies on complex optimized kernels, and matching even a fraction of their performance is still considered beneficial.

\vspace{-2mm}
\section{KernelBench Baseline Evaluation}
\label{sec-4:baseline}
In this section, we investigate how a range of LMs perform when evaluated off-the-shelf on KernelBench and explore their capabilities and failure modes.

\subsection{One-shot Baseline} \label{4.1}

\begin{figure}[H]
    \centering
    \begin{minipage}{0.48\textwidth}
        \centering
        \setlength{\tabcolsep}{2pt} % Reduce space in table
        \begin{tabular}{lccc|ccc}
           \toprule
            $\textbf{\fast{1}}$ over: & \multicolumn{3}{c}{PyTorch Eager} & \multicolumn{3}{c}{torch.compile} \\ 
            \cmidrule(lr){2-4} \cmidrule(lr){5-7}
            KernelBench Level & 1 & 2 & 3 & 1 & 2 & 3 \\ 
            \midrule
            \small{GPT-4o}              & 4\%  & 5\%  & 0\%  & 18\% & 4\%  & \textbf{4\%} \\
            \small{OpenAI o1}           & \underline{10}\% & \underline{24}\% & \textbf{12\%} & 28\% & \underline{19}\% & \textbf{4\%} \\
            \small{DeepSeek V3}         & 6\%  & 4\%  & \underline{8}\%  & 20\% & 2\%  & \underline{2}\%  \\
            \small{DeepSeek R1}         & \textbf{12\%} & \textbf{36\%} & 2\% & \textbf{38\%} & \textbf{37\%} & \underline{2}\%  \\
            \small{Claude 3.5 Sonnet}   & \underline{10}\% & 7\%  & 2\%  & \underline{29}\% & 2\%  & \underline{2}\%  \\
            \small{Llama 3.1-70B Inst.} & 3\%  & 0\%  & 0\%  & 11\% & 0\%  & 0\%  \\
            \small{Llama 3.1-405B Inst.}& 3\%  & 0\%  & 2\%  & 16\% & 0\%  & 0\%  \\
        \bottomrule

        \end{tabular}
        \vspace{3pt} % Adds a little spacing before the caption
        \captionsetup{type=table} % Ensures caption works inside minipage
        \caption{\textbf{KernelBench is a challenging benchmark for current LMs}. Here we present \fast{1}, i.e. the percentage of problems where the model-generated kernel is faster than the PyTorch Eager and \texttt{torch.compile} baseline (default configuration) on NVIDIA L40S.}
        \label{table:greedy-baseline}
    \end{minipage}
    \hfill
    \begin{minipage}{0.48\textwidth}
        \centering
        \includegraphics[width=\textwidth]{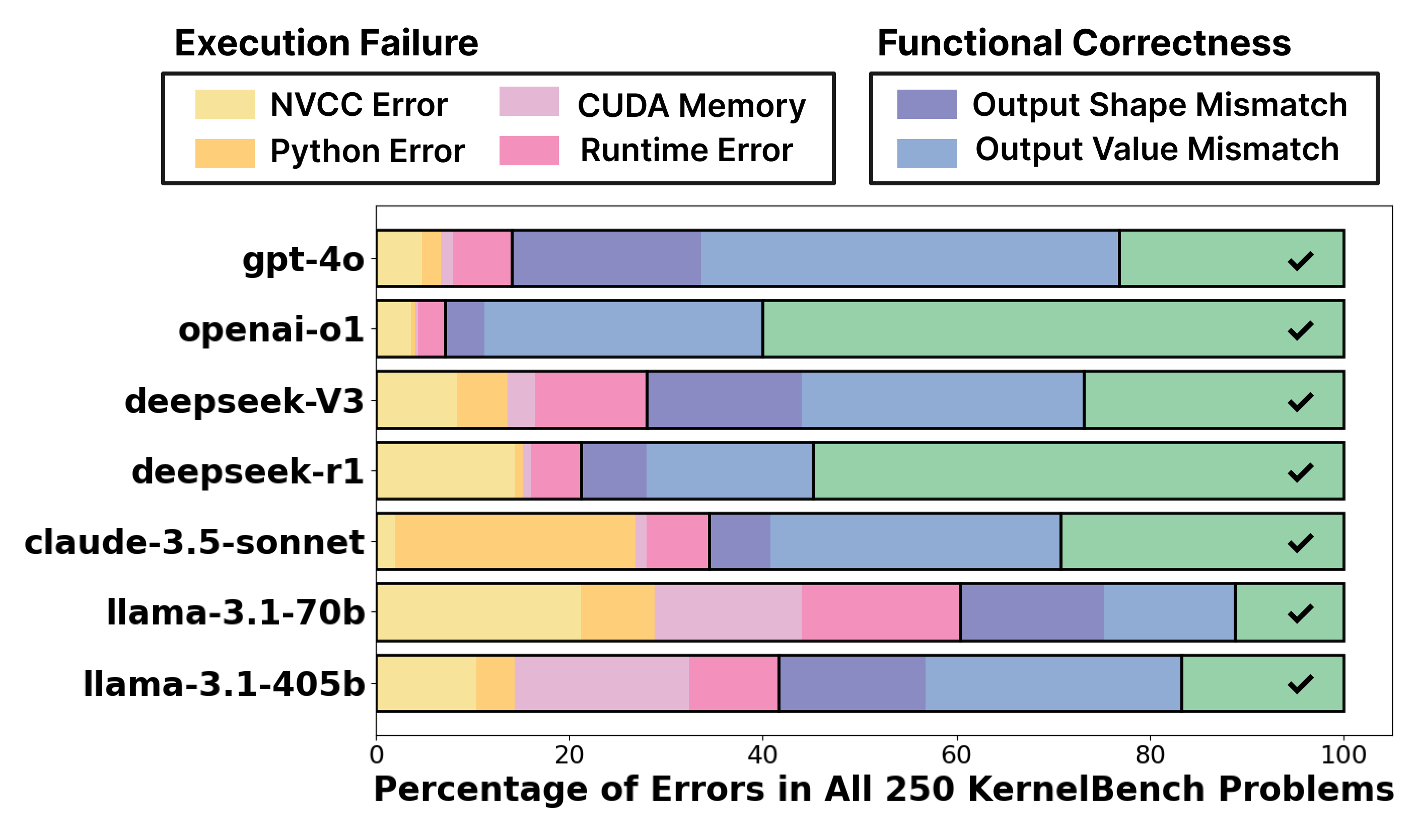} % Adjust width for wrapping
        \caption{\textbf{We categorize failure modes of kernel code into execution failure and functional correctness}. For the one-shot baseline, reasoning models generate fewer kernels with execution failures, but all models struggle similarly with functional correctness.}
        \label{fig:error-breakdown}
    \end{minipage}
    
\end{figure}

We evaluate LMs using a prompt that contains one example of a PyTorch \texttt{Model} input and \texttt{ModelNew} output, highlighting the task format. The example is simple, containing only an \texttt{add} operator
% and an output performing the \texttt{add} in CUDA 
(See Appendix \ref{appendix:one-shot-baseline-prompts}). Given this in-context example and the PyTorch task \texttt{Model} to optimize, the LM generates \texttt{ModelNew} via greedy decoding. We profile the generated code on an NVIDIA L40S GPU, and measure the \fast{p} metric across all problems. Table \ref{table:greedy-baseline} shows that the LM-generated kernels achieves a speedup over PyTorch Eager in fewer than 20\% of tasks on average. 

\footnotetext{The \texttt{torch.compile} baseline runtime is sometimes slower than Torch Eager -- this is due to reproducible runtime overhead (\textit{not compile time}) that could be significant for small kernels in Level 1. We focus on PyTorch Eager for the rest of our analysis, but we elaborate on other baselines in Appendix~\ref{appendix:eval-methods-and-baselines}.}

\subsection{Correctness: Error Analysis} \label{sec-4.2-correctness}
    
In Figure \ref{fig:error-breakdown}, we analyze the failure modes of LMs across problems. It can be seen that a large proportion of model-generated kernels are incorrect. To better understand where model-generated kernels fail, we break down their correctness issues into execution failures (CUDA/\texttt{nvcc} / Python compile-time errors, CUDA memory violations, and runtime errors) and correctness errors (output tensor shape and value mismatches). We observe that the reasoning LMs (o1, R1) produce fewer incorrect solutions ($<55\%$) than other models ($>70\%$). However, we find this is mainly because they make fewer execution failures. All LMs struggle with functional correctness to a similar degree.

\vspace{-2mm}
\subsection{Performance: Speedup Distribution} 
\begin{figure*}
\begin{center}
\vspace{-10pt}
\centerline{\includegraphics[width=\textwidth]{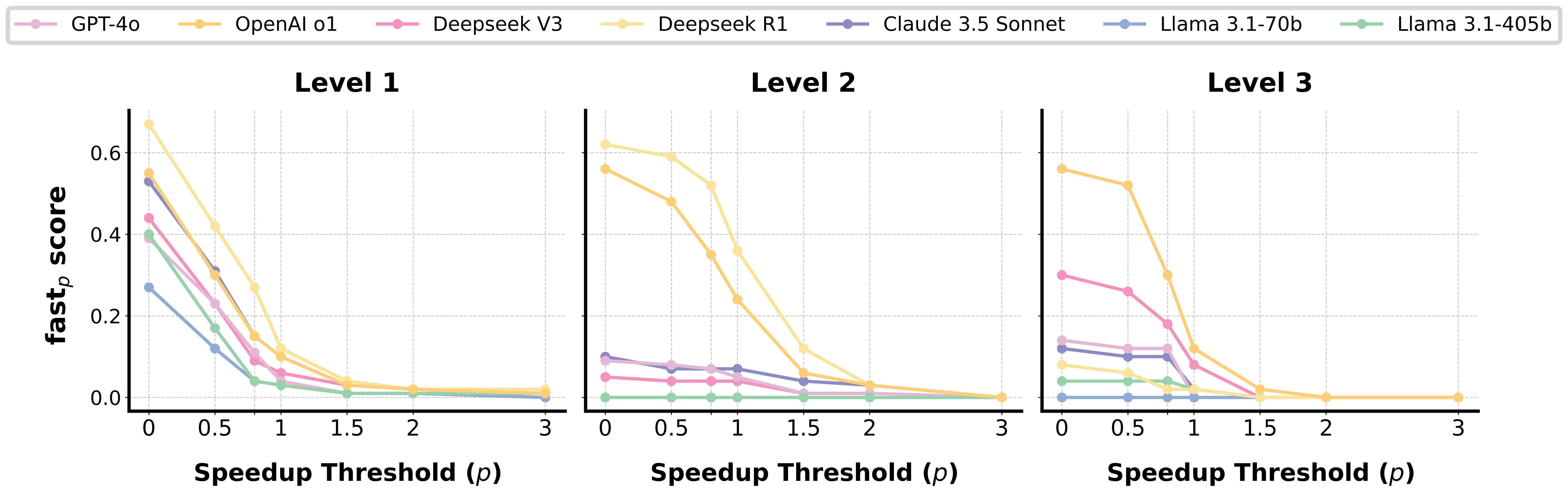}}
\caption{\textbf{Most LM-generated kernels are slow.} This figure shows the distribution of the \fast{p} metric as the speedup threshold $p$ (over PyTorch baseline) increases. \fast{0} represents the number of \textit{correct} kernels regardless of speed, and \fast{1} represents the number of correct kernels achieving at least $> 1\times$ speedup over PyTorch. Increasing the threshold $p$ increases the difficulty.
% as difficulty increases.
% \todo{say something more insightful}
}
\vspace{-20pt}
\label{fig-greedy-fastp}
\end{center}
\end{figure*}

A key point of interest is whether the functionally correct LM-generated kernels outperform the PyTorch baseline. Figure~\ref{fig-greedy-fastp} shows the distribution of \fast{p} as $p$ varies, indicating the percentage of kernels that are $p$-times faster than the PyTorch Eager baseline (the top right of the plot is better). At $p=1$, fewer than 15\% of LM-generated kernels outperform PyTorch across all KernelBench levels. Reasoning-based LMs generally outperform the other LMs in providing speedups.

% \vspace{-1mm}

% (see Appendix \ref{appendix:one-shot-baseline-speedup} for examples). 

\vspace{-2mm}
\subsection{Performance Variations across Hardware}
\label{section:perf-across-hardware}
Our one-shot baseline makes no assumptions about the underlying hardware, so a natural question is how our analysis of the LM-generated kernels generalizes across various GPU types. Table \ref{tab:speedup-hardware-comparison} and Figure \ref{fig:r1_eager_level1_hw} show that kernels outperforming PyTorch Eager on NVIDIA L40S in Level 1 achieve similar speedups versus the baselines on other GPUs. However, on problems in Level 2, LMs exhibit larger variations in speedups  across GPUs (Figure \ref{fig:r1_eager_level2_hw}): DeepSeek R1-generated kernels achieve a \fast{1} of 36\% on NVIDIA L40S but 47\% on NVIDIA A10G for Level 2. This suggests that one-shot LM-generated kernels may not generalize well across hardware.
% , especially as the optimization complexity increases from having a larger optimization space. 
To generate target-specific kernels, we explore in Section \ref{subsection:hardware-info-case-study} whether providing hardware-specific details in-context could help.

Our analysis reveals that the best models available today struggle to generate correct kernels that outperform the baseline PyTorch speeds. LM-generated kernels frequently fail due to simple compiler and run-time errors. Furthermore, it is difficult for LMs to write kernels that perform well across hardware platforms given simple instructions.

\vspace{-2mm}
\section{Analysis of Model Capabilities}
\label{sec-5:analysis}
In the last section, we found that KernelBench is a challenging benchmark for today's models. In this section, we conduct case studies to explore opportunities for improvement in future models and AI systems.

\subsection{Case Study: Leveraging the KernelBench Environment Feedback at Test-Time}
As observed in Section~\ref{sec-4.2-correctness},  execution failures are the most frequent failure mode in LM-generated kernels. The environment provided by KernelBench allows us to collect rich signals, including compiler errors, correctness checks, and runtime profiling metrics, all of which can be fed back in to the LM to help it resolve kernel failures. To explore how well LMs can use this feedback, we evaluate and compare two baselines: (1) generating multiple parallel samples from the LM per KernelBench task and (2) sequentially generating kernels per KernelBench task by allowing the LM to iteratively refine using the execution feedback.

\subsubsection{Repeated Sampling} 
The KernelBench environment enables programmatic verification of LM-generated kernels, 
allowing us to collect and evaluate multiple LM generations \textit{per task} ~\cite{brown2024largelanguagemonkeysscaling,Li_2022,grubisic2024prioritysamplinglargelanguage}. We evaluate this \textit{repeated sampling} approach using $\text{fast}_{p}@k$, which measures the percentage of tasks where the model generated \textbf{at least one functionally correct kernel that is $p$ times faster than PyTorch Eager when drawing $k$ samples}.

\begin{figure}[ht]
    \centering
    \begin{minipage}{0.43\textwidth}
    \textbf{Repeated sampling helps LMs discover more fast and correct solutions.} Figure \ref{fig-multisample-passk} shows that repeated sampling with high temperature improves \fast{1} as $k$ increases across all three levels with both DeepSeek-V3 and Llama 3.1 70B. Notably, on Level 2, DeepSeek-V3 reaches a \fast{1} of 37\% with $k=100$ samples, compared to just 4\% in the one-shot baseline. Examining the samples, we find that high-temperature sampling helps explore the solution space, increasing the chances of generating error-free kernels with better optimizations. However, if a model has a very low inherent probability of solving a task, simply increasing the sampling budget has limited impact. For example, DeepSeek-V3 was never able to generate any correct solution for a group of 34 convolution variants in Level 1, even when attempting with 100 samples.
    
    \end{minipage}
    \hfill
        \begin{minipage}{0.55\textwidth}
        
    \begin{center}
    \vspace{-2mm}
    \centerline{\includegraphics[width=\columnwidth]{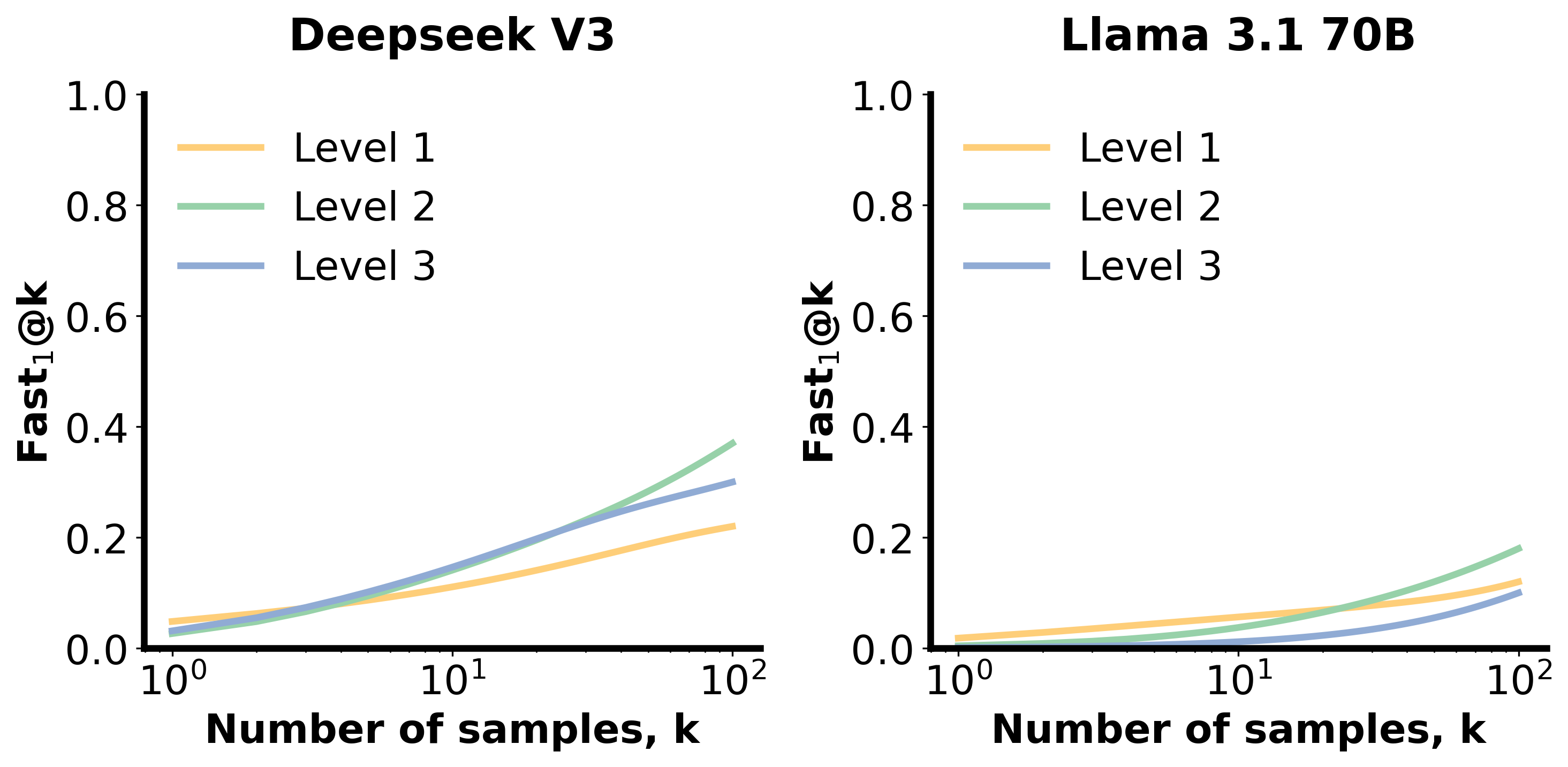}}
    \caption{\textbf{Repeated sampling helps discover more correct and performant kernels}. As the number of repeated samples $k$ increases (up to 100), we observe that $\text{fast}_1$@k improves for both DeepSeek-V3 and Llama 3.1-70B Instruct across all 3 KernelBench levels. We also observe a larger increase in correct solutions for Level 2 kernels. }
    \label{fig-multisample-passk}
    \end{center}

    \end{minipage}
    
\end{figure}

\subsubsection{Iterative Refinement of Generations}
\label{sec:iterative-refinement}

The KernelBench environment is well-suited for collecting compiler feedback, execution errors, and timing analysis using tools like the PyTorch profiler as ground-truth signals. We investigate whether leveraging this feedback can help LMs to iteratively refine their generations.

\begin{figure}[H]
    \centering
    \vspace{-2mm}
    \includegraphics[width=1.0\linewidth]{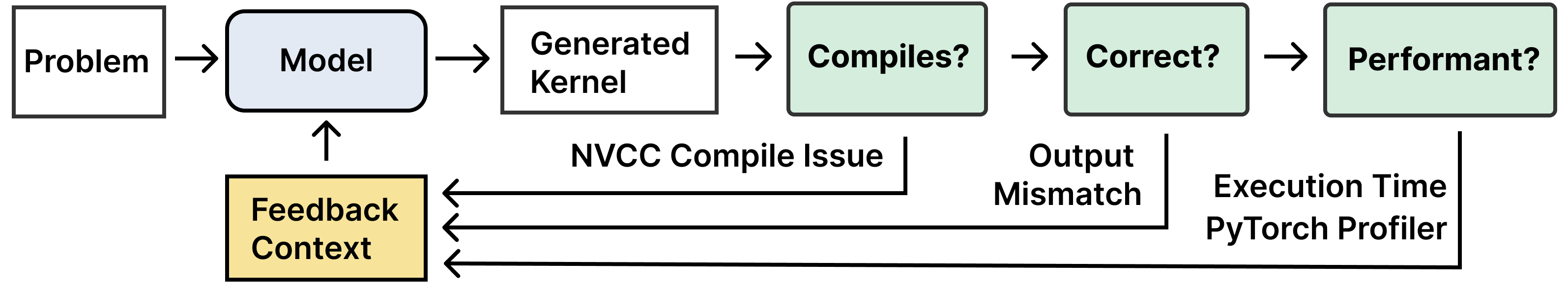}
    \caption{\textbf{The KernelBench framework enables models to receive and leverage feedback during iterative refinement.} These ground-truth signals include NVCC compiler error messages, execution statistics (e.g. correctness checks and wall clock time), and the PyTorch profiler (operator timing breakdown). }
    \vspace{-2mm}
    \label{fig:multi-turn-feedback}
\end{figure}
 
\noindent We provide feedback to the model after each generation in a multi-turn process: after the initial generation, we provide the model with its previous generation $G$, as well as compiler/execution feedback $E$ and/or profiler output $P$ over its current generation. We define each generation and subsequent feedback as a \textit{turn}, and run this \textbf{Iterative Refinement} process over $N$ turns. For each turn, we measure $\text{fast}_{p}@N$, which is the percentage of tasks where the model generated \textbf{at least one} functionally correct kernel that is $p$ times faster than PyTorch Eager by turn $N$.
\FloatBarrier  
\begin{figure}[ht]
    \centering
    \begin{minipage}{0.47\textwidth}
    \textbf{Leveraging execution feedback helps reduce errors and improves overall speedups over time.} We examine the \fast{1} behavior at turn $N=10$ in Table \ref{table:speedup-method-comparison} and find that iterative refinement consistently improves performance across models and levels of KernelBench. DeepSeek-R1 on Level 2 results in the most notable improvement, where the combination of execution feedback $E$ and profiler feedback $P$ boosts \fast{1} from $36\%$ to $72\%$ (shown in Figure~\ref{fig-multiturn-increase-k}). 
    \\\\
    Furthermore, by examining iterative refinement trajectories, we find that models self-correct more effectively with execution feedback $E$, fixing issues especially related to execution errors. DeepSeek-R1 on Level 1 and 2 can generate a functional kernel on \textgreater 90\% of the tasks within $10$ turns of refinement (Table \ref{table:fast0-method-comparison}). However, the remaining incorrect kernels almost always fail due to functional incorrectness, likely because correctness feedback is less granular than execution failure messages. We include successful and failed examples of iterative refinement trajectories in Appendix~\ref{appendix:iterative-refinement}. 
    
    \end{minipage}
    \hfill
    \begin{minipage}{0.49\textwidth}
        
       \centering
        \vspace{-2mm}
        \includegraphics[width=\textwidth]{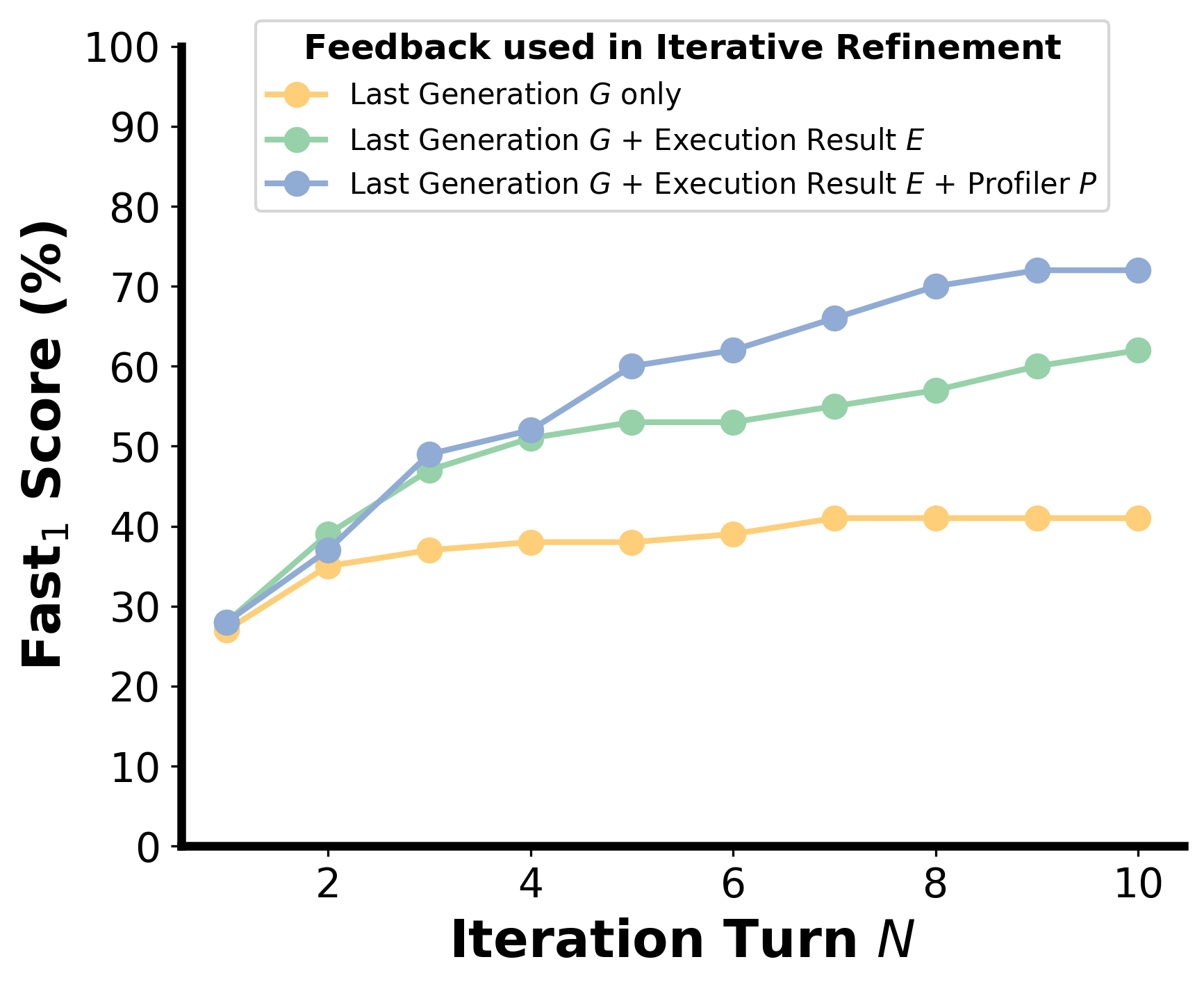} % Slightly reduced width for better wrapping
        \caption{\textbf{Iterative refinement with execution feedback $E$ and profiling information $P$ enable models to improve kernel generations over turns}, as shown in the $\text{fast}_{1}@N$ trajectory of DeepSeek-R1 on Level 2. The percentage of problems where the best generated kernel up to turn $N$ is correct and faster than PyTorch Eager consistently increases with the number of turns.}
        \vspace{-2mm}
        \label{fig-multiturn-increase-k}

    \end{minipage}
    
\end{figure}

\FloatBarrier  

\subsubsection{Comparing Repeated Sampling and Iterative Refinement}
\begin{table*}[ht]
\vspace{-2mm}
\centering
\setlength{\tabcolsep}{2pt} % Reduce column spacing
{\small % Only the table content is affected

\begin{tabular}{l|ccc|ccc|ccc}
\toprule
\multirow{3}{*}{\textbf{Method}} & \multicolumn{3}{c|}{\textbf{Level 1}} & \multicolumn{3}{c|}{\textbf{Level 2}} & \multicolumn{3}{c}{\textbf{Level 3}} \\
 & \scriptsize{Llama-3.1} & \scriptsize{DeepSeek} & \scriptsize{Deepseek} & \scriptsize{Llama-3.1} &  \scriptsize{Deepseek} &  \scriptsize{Deepseek} &  \scriptsize{Llama-3.1} &  \scriptsize{Deepseek} &  \scriptsize{Deepseek} \\
 & \small{70B} & \small{V3} & \small{R1} & \small{70B} & \small{V3} & \small{R1} & \small{70B} & \small{V3} & \small{R1} \\
\midrule
Single Attempt (Baseline) & 3\% & 6\% & 12\% & 0\% & 4\% & 36\% & 0\% & 8\% & 2\% \\
\midrule
Repeated Sampling (@10) & 5\% & 11\% & N/A & 3\% & \textbf{14\%} & N/A & 1\% & 14\% & N/A \\
\midrule
Iterative Refinement w G & \textbf{9\%} & 9\% & 18\% & 0\% & 7\% & 44\% & 0\% & 14\% & 4\% \\
Iterative Refinement w G+E & 5\% & 13\% & 41\% & \textbf{5\%} & 5\% & 62\% & \textbf{8\%} & \textbf{22\%} & 12\% \\
Iterative Refinement w G+E+P & 7\% & \textbf{19\%} & \textbf{43\%} & 4\% & 6\% & \textbf{72\%} & 2\% & 14\% & \textbf{18\%} \\
\bottomrule
\end{tabular}
}
\caption{\textbf{Both repeated sampling and iterative improvement enable models to generate more correct and fast kernels compared to baseline:} Here we present the percentage of problems where the LM-generated kernel is correct and faster than baseline Torch Eager ($\text{Fast}_1$ in \%) for the two test-time methods, both with the same sample budget of $10$ calls. We further compare performance within iterative refinement achieved when leveraging previous Generation $G$, Execution Result $E$, and Timing Profiles $P$. Note we do not repeatedly sample DeepSeek R1, as its API endpoint does not provide a temperature parameter.}
\vspace{-2mm}
\label{table:speedup-method-comparison}
\end{table*}

In Table~\ref{table:speedup-method-comparison}, we compare repeated sampling and iterative refinement given a fixed budget of $10$ inference calls. Both methods provide meaningful improvements over the one-shot baseline, with iterative refinement being more effective in 5 of the 6 cases. However, ultimately we find that the effectiveness of the test-time methods is inherently dependent on the quality of the base model. For instance, with repeated sampling, DeepSeek-V3 consistently outperforms Llama-3.1 70B across all three levels. Similarly, with iterative refinement, DeepSeek-R1 consistently improves using feedback $E$ and $P$, while DeepSeek-V3 and Llama-3.1 70B does not always benefit from having such information. 

\vspace{-2mm}
\subsection{Case Study: Generating Hardware-Efficient Kernels via Hardware Knowledge}
\label{subsection:hardware-info-case-study}

It is clear that LMs demonstrate limited success at generating hardware-efficient kernels. This is likely due to the scarcity of kernel code in the training data and the fact that the optimal kernel may need to change depending on the hardware platform-specific properties, as discussed in Section~\ref{section:perf-across-hardware}. In this case study, we explore providing 1) in-context examples of best-practices for kernel engineering and 2) in-context hardware specification details.

\subsubsection{Hardware-aware In-Context Examples}
\label{subsection:few-shot}

Well-written kernels often use techniques such as fusion, tiling, recompute, and asynchrony to maximize performance. We find that most of the one-shot generated kernels evaluated in \Cref{sec-4:baseline} often do not use these techniques. Here, we explore whether providing explicit in-context examples that use these techniques can help the LMs improve their performance on KernelBench. Specifically, we include three in-context examples: GeLU~\cite{hendrycks2023gaussianerrorlinearunits} using operator fusion, matrix multiplication using tiling ~\cite{mills2024cuda}, and a minimal Flash-Attention~\cite{dao2022flashattention, kim2024flashattention} kernel that demonstrates shared memory I/O management.
\\\\
\noindent \textbf{In-context examples degrade the LM's \textit{overall} \fast{1} score since LMs attempt more aggressive optimization strategies, but result in more execution failures.} OpenAI o1's generations are 25\% longer on average using the few-shot examples, compared to the generations produced by \Cref{sec-4:baseline} baseline. However, among the correct solutions, the LMs apply interesting optimizations: we find that on 77\% of GEMM variants in KernelBench Level 1, o1 applies tiling and improves speed over the one-shot baseline (although remains slower than PyTorch Eager due to the lack of tensor core utilization). On Level 2, o1 applies aggressive shared memory I/O management on 11 problems, and is able to outperform PyTorch Eager (See Appendix~\ref{appendix:few-shot-study}).

\subsubsection{Specifying Hardware Information} 
\label{subsection:hardware-prompting-study}
As discussed in Section~\ref{section:perf-across-hardware}, kernel performance varies depending on the hardware platform. For instance, FlashAttention-2~\cite{dao2023flashattention2} degrades 47\% in hardware utilization going from the NVIDIA A100 to H100 GPU. FlashAttention-3~\cite{dao2024flashattention3}, an entirely different algorithm, was written for the H100. In this study, we explore whether LMs can use (1) hardware specifications such as the GPU type (H100, A100, etc.), memory sizes, bandwidths, TFLOPS and (2) hardware knowledge (e.g. definitions of threads, warps, thread-blocks, streaming multiprocessors) in-context to generate improved kernels (See Appendix~\ref{appendix:cross-hardware-study} for more detail on the context). 
\\\\
\noindent \textbf{Models rarely generate kernels that are optimized for the underlying hardware, highlighting room for improvement for future models.} Certain generations of GPUs (e.g. H100) feature a variety of new hardware units and instructions from their predecessors. Providing hardware information does not significantly impact the outputs of Llama 3.1 70B or DeepSeek-V3. 

Interestingly, we find that a subset of OpenAI o1 and DeepSeek-R1 generated kernels use hardware-specific instructions and optimizations. R1 attempts to generate warp matrix multiply-accumulate (\texttt{wmma}) instructions (Figure \ref{fig:example_r1_generated_kernel_hw}) for approximately $50\%$ of the Level 1 matrix multiplication problems, although most fail to compile. Among the functionally correct generations, R1 and o1 produce 1-3 outliers per level that are $\geq 2\times$ faster than the \Cref{sec-4:baseline} baselines. Overall, we find that LMs are better at adjusting their approaches when provided with few-shot examples in \Cref{subsection:few-shot} than with hardware information.

\vspace{-2mm}
\section{Discussion}
\label{sec-6:discussion}
% In this section, we discuss qualitative examples of LM generations, and discuss opportunities for improvement. 

\vspace{-2mm}
\subsection{Deep Dive Into Interesting Kernels} \label{section: discussion-interesting-kernel}
Here, we discuss a few surprising LM-generated kernels that demonstrate significant speedups over the PyTorch baseline. See detailed examples in Appendix \ref{appendix:kernel-case-study}.
\\\\
\noindent \textbf{Operator fusion} GPUs have small amounts of fast-access memory and large amounts of slow-access memory. Fusion can help reduce slow-access I/O costs by performing multiple operations on data that has been loaded into fast-access memory.
% can reduces I/O costs by eliminating intermediate writes and reads, keeping computations in faster on-chip memory.  
We find that LMs optimize the GELU (2.9x) and Softsign (1.3x) operators by fusing their computations into a single kernel. LMs generated a kernel that fuses multiple foundational operators -- matrix multiplication with division, summation, and scaling -- giving a 2.6x speedup. Overall, LMs leave  many fusion opportunities on the table.
% is, however, a lot more opportunities for fusion (most level 2 problems) that the model is not currently doing.
\\\\
\noindent \textbf{Memory hierarchy} Effective kernels explicitly manage utilization of the limited amounts of shared and register memory. In the generated kernels, we found kernels that uses GPU shared memory -- cosine similarity (2.8x) and triplet margin loss (2.0x) -- to achieve speedups. We did not find successful usages of tensor core instructions, which are crucial for AI performance.
\\\\
\noindent \textbf{Algorithmic optimizations} Kernels can require algorithmic modifications to better utilize the hardware features. We found one interesting generation for the problem of performing a multiplication between a dense and diagonal matrix, where the kernel scales each row (or column), rather than loading the zero-entries of the diagonal matrix, yielding a 13x speedup over PyTorch Eager. 

\subsection{Conclusion}
Our contributions are: (1) We present KernelBench, a framework that lays the groundwork for LM-driven kernel optimization, and (2) We evaluate a diverse set of models and approaches, analyzing their strengths and limitations, and providing insights into opportunities for improvement.

Overall, while most benchmarks eventually saturate, KernelBench is designed to dynamically evolve as new AI workloads arise.  Our \fast{p} metric can be adapted over time to measure the speedup threshold ($p$) over increasingly advanced baselines (i.e., beyond the PyTorch baseline used in our work). Since PyTorch is cross-hardware platform compatible, the PyTorch-based tasks in KernelBench tasks can be evaluated on every \textit{new hardware platform} release. 
Finally, unlike many benchmarks, success on KernelBench directly maps to production value and real-world impacts (lowering costs and reducing energy consumption at scale). These properties ensure that KernelBench will remain valuable in the ever-evolving AI landscape.

\subsection{Opportunities for Future Work}
We show that there is significant room for improvement on KernelBench given the currently available models. 
First, future work can explore the development of advanced fine-tuning and reasoning techniques, including agentic workflows. Since CUDA is a low-resource language, it would be valuable for future work to open-source more high quality data. Second, LMs generate raw CUDA code in our experiments. However, future work can explore whether generating code using alternative programming abstractions (e.g., provided in ThunderKittens, CUTLASS, Triton, and others) can simplify the generation problem, for instance by making it easier for LMs to leverage tensor core instructions. Third, our evaluation has also been limited to GPUs so far and future work can expand to other hardware accelerators. 

\section*{Ethics Statement}
Optimized GPU kernels can lead to significant energy savings in large-scale machine learning workloads, reducing both computational costs and environmental impact. By providing a framework for AI-assisted performance tuning, KernelBench contributes to more energy-efficient AI systems, aligning with global efforts to reduce the carbon footprint of computing infrastructure.

KernelBench does not involve human studies or collect user data, eliminating privacy concerns. It also avoids proprietary or private code, relying solely on publicly available Github repositories.

\section*{Acknowledgements}
We are grateful to Google DeepMind, Google, IBM, Stanford HAI, PrimeIntellect, and Modal for supporting this work. We thank Aaryan Singhal, AJ Root, Allen Nie, Anjiang Wei, Benjamin Spector, Bilal Khan, Bradley Brown, Dylan Patel, Genghan Zhang, Hieu Pham, Hugh Leather, John Yang, Jon Saad-Falcon, Jordan Juravsky, Marcel Rød, Mark Saroufim, Michael Zhang, Minkai Xu, Ryan Ehrlich, Sahan Paliskara, Sahil Jain, Shicheng (George) Liu, Simran Arora, Suhas Kotha, Vikram Sharma Mailthody, and Yangjun Ruan for insightful discussions and constructive feedback in shaping this work.


% TODO: AVOID SUPER LONG CITATION AUTHOR LISTS!
\bibliography{kernelbench}
\bibliographystyle{arxiv}

%%%%%%%%%%%%%%%%%%%%%%%%%%%%%%%%%%%%%%%%%%%%%%%%%%%%%%%%%%%%%%%%%%%%%%%%%%%%%%%
%%%%%%%%%%%%%%%%%%%%%%%%%%%%%%%%%%%%%%%%%%%%%%%%%%%%%%%%%%%%%%%%%%%%%%%%%%%%%%%
% APPENDIX
%%%%%%%%%%%%%%%%%%%%%%%%%%%%%%%%%%%%%%%%%%%%%%%%%%%%%%%%%%%%%%%%%%%%%%%%%%%%%%%
%%%%%%%%%%%%%%%%%%%%%%%%%%%%%%%%%%%%%%%%%%%%%%%%%%%%%%%%%%%%%%%%%%%%%%%%%%%%%%%
\newpage
\appendix

\onecolumn
\section{KernelBench Task Example}
\label{appendix:kernelbench-task-examples}
Here we provide an example task from KernelBench. Each task is wrapped in a class named \texttt{Model}. A task contains two key functions in the \texttt{Model} class, \texttt{\_\_init\_\_} and \texttt{forward}; helper functions are included if necessary. We fix the shape of inputs and vary the numerical values through randomly generated tensors. We provide two functions, \texttt{get\_inputs} and \texttt{get\_init\_inputs}, for generating random parameters for initializing the model and running a forward pass, respectively.
\\\\
\noindent \textbf{PyTorch Reference Architecture:}
\begin{lstlisting}
import torch
import torch.nn as nn

class Model(nn.Module):
    """
    Simple model that performs a single matrix multiplication (C = A * B) with a large K dimension
    """
    def __init__(self):
        super(Model, self).__init__()
    
    def forward(self, A: torch.Tensor, B: torch.Tensor) -> torch.Tensor:
        """
        Performs matrix multiplication of A and B.

        Args:
            A: Input tensor of shape (M, K)
            B: Input tensor of shape (K, N)

        Returns:
            Output tensor of shape (M, N)
        """
        return torch.matmul(A, B)

M = 256
N = 256
K = 131072

def get_inputs():
    A = torch.randn(M, K)
    B = torch.randn(K, N)
    return [A, B]

def get_init_inputs():
    return []  # No special initialization inputs needed
\end{lstlisting}

\noindent \textbf{We include an example output from a model below that attempts to optimize the task specification above.} The model is not only responsible for generating the kernel code, but is also responsible for generating the peripheral code to integrate the kernel into the PyTorch framework. The evaluation framework will run the forward pass of the model as if it is a PyTorch operator, so a common approach is to inline the CUDA code. 
\clearpage
\begin{lstlisting}
import torch
import torch.nn as nn
from torch.utils.cpp_extension import load_inline

# Define the custom CUDA kernel for matrix multiplication
matmul_source = """
#include <torch/extension.h>
#include <cuda_runtime.h>

__global__ void matmul_kernel(const float* A, const float* B, float* C, int M, int N, int K) {
    int row = blockIdx.y * blockDim.y + threadIdx.y;
    int col = blockIdx.x * blockDim.x + threadIdx.x;

    if (row < M && col < N) {
        float sum = 0.0f;
        for (int k = 0; k < K; ++k) {
            sum += A[row * K + k] * B[k * N + col];
        }
        C[row * N + col] = sum;
    }
}

torch::Tensor matmul_cuda(torch::Tensor A, torch::Tensor B) {
    int M = A.size(0);
    int N = B.size(1);
    int K = A.size(1);

    auto C = torch::zeros({M, N}, A.options());

    dim3 threadsPerBlock(16, 16);
    dim3 numBlocks((N + threadsPerBlock.x - 1) / threadsPerBlock.x, (M + threadsPerBlock.y - 1) / threadsPerBlock.y);

    matmul_kernel<<<numBlocks, threadsPerBlock>>>(A.data_ptr<float>(), B.data_ptr<float>(), C.data_ptr<float>(), M, N, K);

    return C;
}
"""

matmul_cpp_source = (
    "torch::Tensor matmul_cuda(torch::Tensor A, torch::Tensor B);"
)

# Compile the inline CUDA code for matrix multiplication
matmul = load_inline(
    name="matmul",
    cpp_sources=matmul_cpp_source,
    cuda_sources=matmul_source,
    functions=["matmul_cuda"],
    verbose=True,
    extra_cflags=[""],
    extra_ldflags=[""],
)


class ModelNew(nn.Module):
    def __init__(self):
        super(ModelNew, self).__init__()
        self.matmul = matmul

    def forward(self, A: torch.Tensor, B: torch.Tensor) -> torch.Tensor:
        return self.matmul.matmul_cuda(A, B)
\end{lstlisting}

\section{Evaluation Methodology and Baselines}
\label{appendix:eval-methods-and-baselines}
All evaluations are conducted on a bare-metal NVIDIA L40S GPU with Ada Lovelace architecture unless otherwise stated (such as the device generalization experiments in Section~\ref{section:perf-across-hardware} and the hardware case study in \ref{subsection:hardware-info-case-study}). The NVIDIA L40S has 48 GB of HBM memory and operates at 300W. Our environment uses Python \texttt{3.10}, PyTorch \texttt{2.5.0+cu124}, and CUDA \texttt{12.4}, which is also where our PyTorch Eager and \texttt{torch.compile} baselines are derived from.

\subsection{Kernel Evaluation Setup}
Recall the KernelBench task entails a PyTorch reference module \texttt{Model} as baseline, and model-generated PyTorch architecture \texttt{ModelNew} with custom inline CUDA kernel. 
\\\\
\noindent For \textbf{correctness}, we set \textit{num\_correctness} to 5, where we check equivalence of output between reference architecture \texttt{Model} and generated architecture with custom kernel \texttt{ModelNew} with 5 randomized inputs. We elaborate on our choice in Appendix~\ref{append:correctness-vary}.
\\\\
\noindent For \textbf{performance}, we measure the wall-clock execution time of \texttt{nn.module.forward} for both \texttt{Model} and \texttt{ModelNew}.  We ensure only one kernel is being evaluated (no other CUDA process) on current GPU. We warm up for 3 iterations and then set  \textit{num\_profile} to 100 times which measures the elapsed execution time signaled between CUDA events \texttt{torch.cuda.Event}.  We take the mean of the 100 trials, and also note its max, min, and standard deviation. While the wall clock time might vary for every trial, we note our coefficient of variation (CV): $\text{std} / \text{mean}$ is consistently $<3\%$, we use the mean of both measured wall clock time for comparisons. 

To compute the speedup of generated architecture over baseline architecture for individual problems, we use the mean for both $\text{speedup} = T_{Model} / 
 T_{ModelNew}$. For example, if $T_{Model}=2$ ms and $T_{ModelNew}=1$ ms, we have a 2x speedup with the newly generated kernel. We compare this speedup with our speedup threshold parameter $p$ (as explained in section \ref{sec-metric-design}) to compute \fast{p} scores. 

\subsection{Correctness Analysis Varying Number of Randomly Generated Inputs}
\label{append:correctness-vary}
Checking equivalence of programs in a formal sense is undecidable. "The Halting Problem" \citep{turing1936a} states that it is impossible to decide, in general, whether a given program will terminate for every possible input. This problem naturally extends to checking equivalence because in order to check whether two programs are equivalent, it is necessary to check their behavior for all inputs, including cases where one or both programs may not terminate. Since determining whether a program halts on a given input is undecidable (the Halting Problem), checking equivalence also becomes undecidable.

Approximate or heuristic methods are often used in practice for checking program equivalence. Random testing is the most common practical approach, where the program is run with sets of randomly chosen inputs, and their outputs are compared. Random testing is particularly effective for AI kernels, where control flow is simpler and the focus is primarily on numerical correctness. By using diverse inputs, it can uncover errors in computations or memory handling with high probability. Evaluating correctness more systematically, especially in the presence of subtle hardware-specific behavior, is an area for further exploration. Future work could investigate formal verification tools to provide stronger guarantees of equivalence.

We use five sets of random inputs for correctness, which is a good tradeoff between the ability to catch errors and efficiency. In an experiment with 100 generated kernels, the results were as follows: 50 kernels were correct (all 5/5 and 100/100), 19 had output value mismatches (19 0/5 and 0/100), 4 had output shape mismatches, 10 encountered runtime errors, and 17 had compilation errors. Notably, the 0/5 and 0/100 failures indicate that no partial correctness was observed.

\subsection{Distribution of Model Performance for One-Shot Baseline} \label{appendix:one-shot-baseline-speedup}
\begin{figure}[H]
    \centering
    \includegraphics[width=1.0\linewidth]{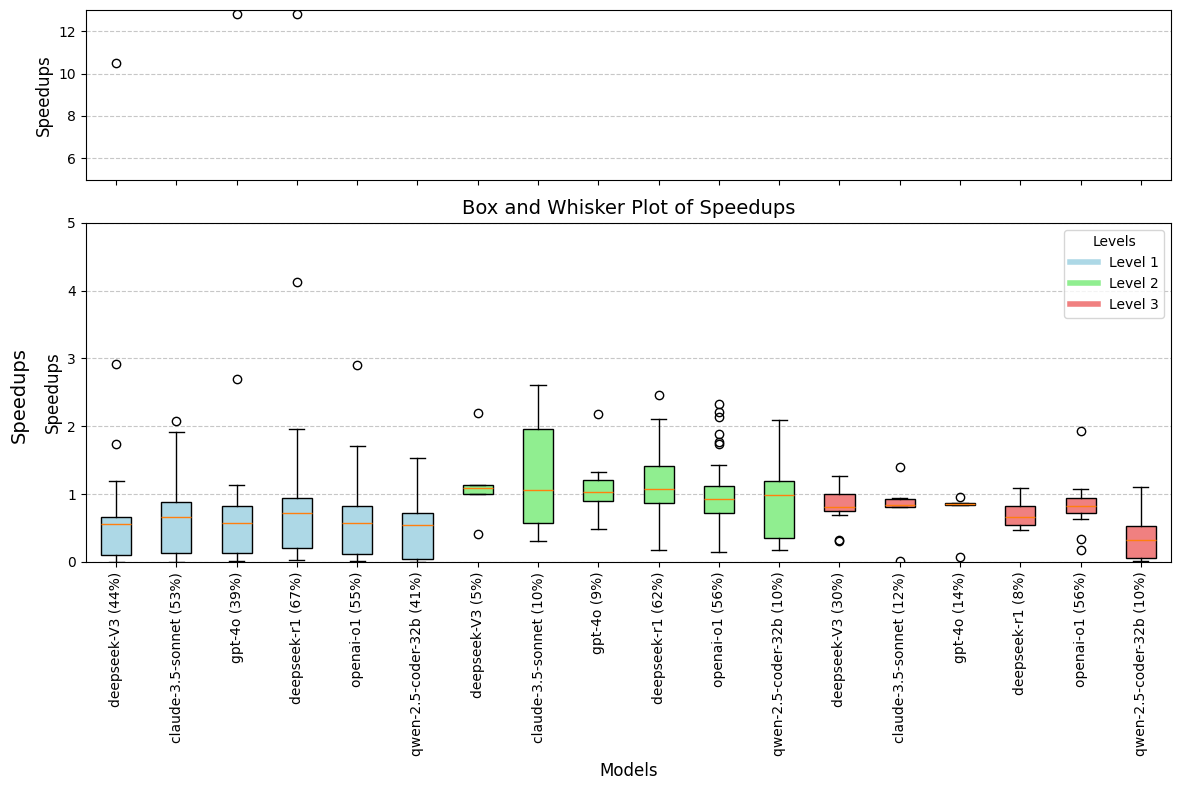}
    \caption{A box and whisker plot of the speedup relative to Torch Eager of (correct) kernels generated by various models in the one-shot baseline setting. We also write the percentage of correctly generated kernels next to the model name. We observe that among most models, the median speedup for correctly generated kernels is below 1.}
    \label{fig-greedy-speedup-box-whiskers}
\end{figure}

Here we examine the quality of (functionally correct) kernel generations across a wide variety of models. Figure~\ref{fig-greedy-speedup-box-whiskers} shows the distribution of speedups for various kernels across different levels and models. The median speedup for both Level 1 and Level 3 are less than 1, and the median speedup for Level 2 is only slightly above one. Level 1 has the most significant outliers, in one case showing a speedup greater than 10. We explored some of these outlier cases in greater detail in Section~\ref{sec-6:discussion}.
\\\\
\noindent \textbf{Reasoning-optimized models (OpenAI-o1 and DeepSeek-R1) perform the best of out-of-the-box across all levels.} These models demonstrate superior kernel generation capabilities, particularly excelling at Level 2 tasks (which mainly involve kernel fusion). In contrast, Llama 3.1 models (both 405B and 70B) perform poorly regardless of model size, suggesting that larger models do not necessarily guarantee better results for this task. DeepSeek-R1, while strong at Level 1 and 2, suffers significantly at Level 3, often generating incorrect kernels.

\subsection{PyTorch Baselines}
PyTorch offers two common execution modes: Eager and \texttt{torch.compile}. 
Aside from the results shown in Table~\ref{table:greedy-baseline}, all performance analysis is evaluated against PyTorch Eager. 
\\\\
\noindent \textbf{PyTorch Eager} is the default execution mode of PyTorch, which dynamically executes computation by invoking calls to highly optimized closed-source kernels. 
\\\\
\noindent \textbf{PyTorch Compile} or \texttt{torch.compile} uses rule-based heuristics over the underlying computation graph during an initial compilation phase and invokes various backends to perform optimizations like kernel fusion and graph transformations. In Table~\ref{table:greedy-baseline}, our performance baseline for \texttt{torch.compile} assumes the default configuration using PyTorch Inductor in default mode. Furthermore, we \textbf{exclude} the \texttt{torch.compile} compile time in our timing analysis, as we are only interested in the raw runtime behavior. \texttt{torch.compile} features multiple other backends and configurations, which we describe in Table \ref{tab:pytorch_configurations}. 

We observe that the \texttt{torch.compile} baseline runtime is generally faster on Level 2 and 3 of KernelBench reference problems compared to PyTorch Eager, mostly due to the availability of graph-level optimizations like operator fusion. However, on Level 1 problems, \texttt{torch.compile} can exhibit higher runtimes than PyTorch Eager, which can be attribute to empirically-reproducible runtime overhead for \texttt{torch.compile} (\textit{not compile time}) that is significant for small kernels.

\begin{table}[ht]
\centering
\begin{tabular}{llll}

\hline
\textbf{Configuration} & \textbf{Backend} & \textbf{Mode} & \textbf{Description} \\ 
\hline
PyTorch (Eager)          & -                & -             & Standard PyTorch eager execution \\ 
Torch Compile          & inductor         & default       & Default \texttt{torch.compile} behavior \\ 
Torch Compile          & inductor         & reduce-overhead & Optimized for reduced overhead \\ 
Torch Compile          & inductor         & max-autotune  & Max autotuning enabled \\ 
Torch Compile          & inductor         & max-autotune-no-cudagraphs & Max autotuning without CUDA graphs \\ 
Torch Compile          & cudagraphs       & -             & CUDA graphs with AOT Autograd \\ 
\hline
\end{tabular}
\caption{Configurations and modes for PyTorch execution and optimization backends.}
\label{tab:pytorch_configurations}
\end{table}

% Greey baseline table
\noindent \textbf{Other \texttt{torch.compile} backends}. In Table~\ref{table:greedy-baseline-compile}, we show more one-shot baseline results for \fast{1} against some of the other \texttt{torch.compile} baselines. We note on some other configurations \fast{1} drops especially for Level 2, as the \texttt{torch.compile} backends apply more aggressive optimization (at the cost of extra compile-time overhead, which we do not measure). Due to the variability of \texttt{torch.compile} across configurations, we focus our analysis on PyTorch Eager. 

\begin{table}[h]
\centering
\setlength{\tabcolsep}{2pt} % default is about 6pt
\begin{tabular}{lccc|ccc|ccc|ccc|ccc}
    \toprule
    \fast{1} over: & \multicolumn{3}{c}{\makecell{torch.compile \\ default}} & \multicolumn{3}{c}{cudagraphs} & \multicolumn{3}{c}{max-autotune} & \multicolumn{3}{c}{\makecell{max-autotune\\no-cudagraphs}} & \multicolumn{3}{c}{reduce-overhead} \\
    \cmidrule(lr){2-4} \cmidrule(lr){5-7} \cmidrule(lr){8-10} \cmidrule(lr){11-13} \cmidrule(lr){14-16}
    KernelBench Level & 1 & 2 & 3 & 1 & 2 & 3 & 1 & 2 & 3 & 1 & 2 & 3 & 1 & 2 & 3 \\
    \midrule
    Claude 3.5 Sonnet   & \underline{29\%} & 2\%  & \underline{2\%}  & 31\% & 7\%  & 2\%  & 31\% & 2\% & 0\% & 29\% & 2\% & 2\% & 31\% & 2\% & \underline{0\%}  \\
    DeepSeek V3         & 20\%  & 2\%  & \underline{2\%}  & 21\% & 4\%  & \underline{20\%}  & 21\% & 2\% & \underline{2\%} & 20\% & 2\% & 2\% & 21\% & 2\% & \underline{0\%}  \\
    DeepSeek R1         & \textbf{38\%} & \textbf{37\%} & \underline{2\%} & \textbf{42\%} & \textbf{52\%} & 0\% & \textbf{42\%} & \textbf{29\%} & 0\% & \textbf{38\%} & \textbf{32\%} & \underline{4\%} & \textbf{42\%} & \textbf{28\%} & \underline{0\%}  \\
    GPT-4o              & 18\%  & 4\%  & \textbf{4\%}  & 22\% & 6\%  & 6\%  & 21\% & 4\% & \underline{2\%} & 18\% & 3\% & \underline{4\%} & 21\% & 4\% & \underline{0\%}  \\
    Llama 3.1-70B Inst. & 11\%  & 0\%  & 0\%  & 12\% & 0\%  & 0\%  & 12\% & 0\% & 0\% & 11\% & 0\% & 0\% & 12\% & 0\% & \underline{0\%}  \\
    Llama 3.1-405B Inst.& 16\%  & 0\%  & 0\%  & 16\% & 0\%  & 4\%  & 16\% & 0\% & 0\% & 16\% & 0\% & 0\% & 16\% & 0\% & \underline{0\%}  \\
    OpenAI O1           & 28\% & \underline{19\%}  & \textbf{4\%}  & \underline{33\%} & \underline{37\%}  & \textbf{26\%}  & \underline{34\%} & \underline{8\%} & \textbf{4\%} & \underline{30\%} & \underline{19\%} & \textbf{6\%} & \underline{34\%} & \underline{8\%} & \textbf{2\%}  \\
    \bottomrule
\end{tabular}

\caption{We compare KernelBench  \texttt{torch.compile} baseline runtime across various configurations, all measured on NVIDIA L40S, in addition to what is showed in Table \ref{table:greedy-baseline}.}

\label{table:greedy-baseline-compile}
\end{table}

\section{Experiment Prompting Details}
\label{appendix:experiment-prompts}
We provide details for the prompting strategies and associated sampling strategies used in Section~\ref{sec-4:baseline} and Section~\ref{sec-5:analysis}. 

\subsection{One-shot Baseline Prompt}\label{appendix:one-shot-baseline-prompts}
For the one-shot baseline as shown in Section~\ref{4.1}, we want to examine each model's out-of-the-box ability to generate kernels by providing the minimum set of information while ensuring the instructions and output format are clear. We query each model with the following prompt and a pair of in-context \texttt{add} examples (the PyTorch reference \texttt{add} and its CUDA kernel counterpart using inline compilation) to provide the output format. We sample the model with greedy decoding to ensure deterministic output, which is setting $\text{temperature}=0$.  

\begin{lstlisting}
You write custom CUDA kernels to replace the pytorch operators in the given architecture 
to get speedups. 

You have complete freedom to choose the set of operators you want to replace. You may
make the decision to replace some operators with custom CUDA kernels and leave others
unchanged. You may replace multiple operators with custom implementations, consider
operator fusion opportunities (combining multiple operators into a single kernel, for
example, combining matmul+relu), or algorithmic changes (such as online softmax). You are
only limited by your imagination.

Here\'s an example to show you the syntax of inline embedding custom CUDA operators in 
torch: The example given architecture is:
```
import torch
import torch.nn as nn
import torch.nn.functional as F


class Model(nn.Module):
    def __init__(self) -> None:
        super().__init__()

    def forward(self, a, b):
        return a + b


def get_inputs():
    # randomly generate input tensors based on the model architecture
    a = torch.randn(1, 128).cuda()
    b = torch.randn(1, 128).cuda()
    return [a, b]


def get_init_inputs():
    # randomly generate tensors required for initialization based on the model architecture
    return []
```

The example new arch with custom CUDA kernels looks like this: 
```
import torch
import torch.nn as nn
import torch.nn.functional as F
from torch.utils.cpp_extension import load_inline

# Define the custom CUDA kernel for element-wise addition
elementwise_add_source = """
#include <torch/extension.h>
#include <cuda_runtime.h>

__global__ void elementwise_add_kernel(const float* a, const float* b, float* out, int size) {
    int idx = blockIdx.x * blockDim.x + threadIdx.x;
    if (idx < size) {
        out[idx] = a[idx] + b[idx];
    }
}

torch::Tensor elementwise_add_cuda(torch::Tensor a, torch::Tensor b) {
    auto size = a.numel();
    auto out = torch::zeros_like(a);

    const int block_size = 256;
    const int num_blocks = (size + block_size - 1) / block_size;

    elementwise_add_kernel<<<num_blocks, block_size>>>(a.data_ptr<float>(), b.data_ptr<float>(), out.data_ptr<float>(), size);

    return out;
}
"""

elementwise_add_cpp_source = "torch::Tensor elementwise_add_cuda(torch::Tensor a, torch::Tensor b);"

# Compile the inline CUDA code for element-wise addition
elementwise_add = load_inline(
    name='elementwise_add',
    cpp_sources=elementwise_add_cpp_source,
    cuda_sources=elementwise_add_source,
    functions=['elementwise_add_cuda'],
    verbose=True,
    extra_cflags=[''],
    extra_ldflags=['']
)

class ModelNew(nn.Module):
    def __init__(self) -> None:
        super().__init__()
        self.elementwise_add = elementwise_add

    def forward(self, a, b):
        return self.elementwise_add.elementwise_add_cuda(a, b)
```

You are given the following architecture: 

<PyTorch reference architecture for specific KernelBench Problem>

Optimize the architecture named Model with custom CUDA operators! Name your optimized
output architecture ModelNew. Output the new code in codeblocks. Please generate real
code, NOT pseudocode, make sure the code compiles and is fully functional. Just output
the new model code, no other text, and NO testing code! 
\end{lstlisting}

\subsection{Repeated Sampling Prompts}\label{appendix:multi-sampling-baseline-prompts}
For repeated sampling, we use the same prompt that we used for the one-shot baseline in Appendix~\ref{appendix:one-shot-baseline-prompts}. We used the same sampling temperature described in \cite{brown2024largelanguagemonkeysscaling} as they allow sample diversity while ensuring quality. Specifically we use $\text{temperature}=1.6$ for Deepseek-V3 and  $\text{temperature}=0.7$ for Llama 3.1-70B.

\subsection{Iterative Refinement Prompts}\label{appendix:multi-turn-baseline-prompts}
For iterative refinement, we start with the same initial prompt that we used for the one-shot baseline in Appendix~\ref{appendix:one-shot-baseline-prompts}. A limitation of our experiments is that we sample with temperature$=0$ to focus on the effect of iterating based on feedback rather than introducing variability. On subsequent generations, we prompt the model with the following template depending on the feedback it expects:
\begin{lstlisting}
<Initial prompt from one-shot baseline for specific KernelBench problem.>

Here is your latest generation:
<Previously generated kernel G>

Your generated architecture ModelNew and kernel was evaluated on GPU and checked against the reference architecture Model.
Here is your Evaluation Result:

<Raw Compiler and Execution Feedback from stdout>

<'if correct:'>
Your kernel executed successfully and produced the correct output.
Here is your wall clock time: {runtime} milliseconds 

<Profiler information if used and correct.>

Name your new improved output architecture ModelNew. Output the new code in codeblocks. Please generate real code, NOT pseudocode, make sure the code compiles and is fully functional. Just output the new model code, no other text, and NO testing code!
\end{lstlisting}

\noindent For the compiler and execution feedback, we handle timeouts and deadlocks explicitly with "Your kernel execution timed out", but do not provide any other information.

\subsection{Few-Shot in Context Prompts} \label{appendix:few-shot-study-prompts}
For Few-Shot experiments as outlined in Section \ref{subsection:few-shot}. We provide more details about the in-context example in Appendix \ref{appendix:few-shot-study}. We sampled these experiments with $\text{temperature}=0$.  
\begin{lstlisting}
<Initial Task prompt from one-shot baseline for Instruction>
<Initial pair of Reference PyTorch and CUDA kernel equiavlent for example add kernel from one-shot baseline for Instruction>

Example <i>
Here is an example architecture
<PyTorch reference architecture for No. i in-context example>

Here is an optimized verison with custom CUDA kernels: 
<PyTorch architecture with Custom CUDA Kernel for No. i in-context example>

.. up to number of in-context sample times


Task:
Here is an example architecture:

<PyTorch reference architecture for specific KernelBench Problem>

Name your new improved output architecture ModelNew. Output the new code in codeblocks. Please generate real code, NOT pseudocode, make sure the code compiles and is fully functional. Just output the new model code, no other text, and NO testing code!
\end{lstlisting}

\subsection{Hardware Case Study Prompts} \label{appendix:hardware-case-study-prompts}

Here we provide hardware information. This is used in Section \ref{section:perf-across-hardware} and elaborated more in \ref{appendix:cross-hardware-study}, sampled with $\text{temperature}=0$.  

\begin{lstlisting}
<Initial Task prompt from one-shot baseline for Instruction>
<Initial pair of Reference PyTorch and CUDA kernel equiavlent for example add kernel from one-shot baseline for Instruction>

Here is some information about the underlying hardware that you should keep in mind. 

The GPU that will run the kernel is NVIDIA <GPU NAME>.

- We have <x> GB GDDR6 with ECC of GPU Memory.
- We have <x> GB/s of Memory Bandwidth.
- We have <x> of RT Core Performance TFLOPS.
- We have <x> of FP32 TFLOPS.
- We have <x> of TF32 Tensor Core TFLOPS.
- We have <x> of FP16 Tensor Core TFLOPS.
- We have <x> of FP8 Tensor Core TFLOPS.
- We have <x> of Peak INT8 Tensor TOPS.
- We have <x> of Peak INT4 Tensor TOPS.
- We have <x> 32-bit registers per SM of Register File Size.
- We have <x> of Maximum number of registers per thread.
- We have <x> of Maximum number of thread blocks per SM.
- We have <x> KB of Shared memory capacity per SM.
- We have <x> KB of Maximum shared memory per thread block.



Here are some concepts about the GPU architecture that could be helpful: 

- Thread: A thread is a single execution unit that can run a single instruction at a time.
- Thread Block: A thread block is a group of threads that can cooperate with each other.
- Shared Memory: Shared memory is a memory space that can be accessed by all threads in a thread block.
- Register: A register is a small memory space that can be accessed by a single thread.
- Memory Hierarchy: Memory hierarchy is a pyramid of memory types with different speeds and sizes.
- Memory Bandwidth: Memory bandwidth is the rate at which data can be read from or stored into memory.
- Cache: Cache is a small memory space that stores frequently accessed data.
- HBM: HBM is a high-bandwidth memory technology that uses 3D-stacked DRAM.

Here are some best practices for writing CUDA kernels on GPU

- Find ways to parallelize sequential code.
- Minimize data transfers between the host and the device.
- Adjust kernel launch configuration to maximize device utilization.
- Ensure that global memory accesses are coalesced.
- Minimize redundant accesses to global memory whenever possible.
- Avoid long sequences of diverged execution by threads within the same warp.
  #We added this to reference the specific GPU architecture
- Use specialized instructions based on the specific GPU architecture

You are given the following architecture: 

<PyTorch reference architecture for specific KernelBench Problem>

Name your new improved output architecture ModelNew. Output the new code in codeblocks. Please generate real code, NOT pseudocode, make sure the code compiles and is fully functional. Just output the new model code, no other text, and NO testing code!
\end{lstlisting}

\section{Kernels of Interest}
\label{appendix:kernel-case-study}
In this section we provide examples of interesting or notable kernel generations. We first expand on the discussion in Section~\ref{sec-6:discussion}, where we defined the following categories of optimizations: algorithmic optimizations, operator fusion, and using hardware features. 

\subsection{Algorithmic Optimizations}
\textbf{\textit{13x Speedup} on Level 1 Problem 11 by Claude-3.5 Sonnet} \\
The original torch operator is \texttt{torch.diag(A) @ B}, multiplying a diagonal matrix formed from the vector \texttt{A} with the matrix \texttt{B}. The model identifies an optimization in the special case of a diagonal matrix multiplication, where the diagonal matrix doesn't need to be explicitly constructed. Instead, each element of the vector \texttt{A} is directly multiplied with the corresponding row in matrix \texttt{B}, significantly improving performance:

\begin{figure}[H]
\begin{lstlisting}[basicstyle=\scriptsize\ttfamily]
__global__ void diag_matmul_kernel(
    const float* diag,
    const float* mat,
    float* out,
    const int N,
    const int M) {
    
    const int row = blockIdx.y * blockDim.y + threadIdx.y;
    const int col = blockIdx.x * blockDim.x + threadIdx.x;
    
    if (row < N && col < M) {
        out[row * M + col] = diag[row] * mat[row * M + col];
    }
}
\end{lstlisting}
\end{figure}

\subsection{Kernel Fusion}
\textbf{\textit{2.9x Speedup} on Level 1 Problem 87 by DeepSeek-V3} \\
GeLU reference in torch:

\begin{figure}[H]
\begin{lstlisting}
0.5 * x * (1.0 + torch.tanh(math.sqrt(2.0 / math.pi) * (x + 0.044715 * torch.pow(x, 3.0))))
\end{lstlisting}
\end{figure}

\noindent Optimized version fuses in a single kernel. There is also a small constant folding optimization, instead of computing \texttt{math.sqrt(2.0 / math.pi)} repeatedly, the kernel uses the precomputed value \texttt{0.7978845608028654f}:
\begin{figure}[H]
\begin{lstlisting}
__global__ void gelu_kernel(const float* x, float* out, int size) {
    int idx = blockIdx.x * blockDim.x + threadIdx.x;
    if (idx < size) {
        float x_val = x[idx];
        float cdf = 0.5f * (1.0f + tanhf((0.7978845608028654f * (x_val + 0.044715f * x_val * x_val * x_val))));
        out[idx] = x_val * cdf;
    }
}
\end{lstlisting}
\end{figure}

\noindent \textbf{\textit{1.3x Speedup} on Level 1 Problem 29 by Claude-3.5 Sonnet} \\
SoftSign reference in torch:
\begin{figure}[H]
\begin{lstlisting}
x / (1 + torch.abs(x))
\end{lstlisting}
\end{figure}

Fused kernel:
\begin{figure}[H]
\begin{lstlisting}
__global__ void softsign_kernel(const float* input, float* output, int size) {
    int idx = blockIdx.x * blockDim.x + threadIdx.x;
    if (idx < size) {
        float x = input[idx];
        float abs_x = abs(x);
        output[idx] = x / (1.0f + abs_x);
    }
}
\end{lstlisting}
\end{figure}

\noindent \textbf{\textit{2.6x Speedup} on Level 2 Problem 13 by Claude-3.5 Sonnet} \\
Sequence of operators in torch:
\begin{figure}[H]
\begin{lstlisting}
x = torch.matmul(x, self.weight.T)  # Gemm
x = x / 2  # Divide
x = torch.sum(x, dim=1, keepdim=True) # Sum
x = x * self.scaling_factor  # Scaling
\end{lstlisting}
\end{figure}

\noindent Fused kernel:
\begin{figure}[H]
\begin{lstlisting}
__global__ void fused_ops_kernel(
    const float* input,
    const float* weight,
    float* output,
    const float scaling_factor,
    const int batch_size,
    const int input_size,
    const int hidden_size
) {
    // Each thread handles one element in the batch
    const int batch_idx = blockIdx.x * blockDim.x + threadIdx.x;
    
    if (batch_idx < batch_size) {
        float sum = 0.0f;
        
        // Compute matmul and divide for this batch element
        for(int h = 0; h < hidden_size; h++) {
            float elem = 0.0f;
            for(int i = 0; i < input_size; i++) {
                elem += input[batch_idx * input_size + i] * 
                        weight[h * input_size + i];
            }
            // Divide by 2 as we go
            sum += (elem / 2.0f);
        }
        
        // Scale and store final result
        output[batch_idx] = sum * scaling_factor;
    }
}
\end{lstlisting}
\end{figure}
\noindent Despite this good example, overall Level 2 generated kernels show insufficient fusion. It is expected that most Level 2 problems can be expressed in a single fused kernel. 
\\\\
\noindent \textbf{\textit{1.9x Speedup} on Level 3 Problem 49 by OpenAI-o1} \\
We have a hypothetical architecture of the attention mechanism where the softmax is replaced with a ReLU.
\begin{figure}[H]
\begin{lstlisting}
    att = (q @ k.transpose(-2, -1)) * (1.0 / math.sqrt(k.size(-1)))
    att = att.masked_fill(self.bias[:,:,:T,:T] == 0, float('-inf'))
    att = F.relu(att)
\end{lstlisting}
\end{figure}
\noindent The model found an optimization that fuses the scaling, masked fill, and ReLU but not anything else, resulting in a modest improvement of 1.9x.
\begin{figure}[H]
\begin{lstlisting}
__global__ void fused_masked_fill_scale_relu_kernel(
    const float* __restrict__ att,
    const float* __restrict__ bias,
    float* __restrict__ output,
    int total_elems,
    float scale,
    int T,
    float negative_infinity
) {
    int idx = blockIdx.x * blockDim.x + threadIdx.x;
    if (idx < total_elems) {
        float val = att[idx] * scale;
        int bias_idx = idx % (T * T);
        if (bias[bias_idx] == 0.0f) {
            val = negative_infinity;
        }
        if (val < 0.0f) {
            val = 0.0f;
        }
        output[idx] = val;
    }
}
\end{lstlisting}
\end{figure}

\subsection{Hardware Features}
\textbf{\textit{2.8x Speedup} on Level 1 Problem 96 by OpenAI-o1} \\
Torch reference for Cosine Similarity Loss
\begin{figure}[H]
\begin{lstlisting}
cosine_sim = torch.nn.functional.cosine_similarity(predictions, targets, dim=1)
return torch.mean(1 - cosine_sim)
\end{lstlisting}
\end{figure}
The generated kernel uses shared memory for reduce redundant global memory accesses, improving data locality and increasing overall performance. This is a moderately complicated kernel with synchronization points and reductions that would be tricky for humans to get right.
\begin{figure}[H]
\begin{lstlisting}
__global__ void cosine_similarity_loss_kernel(
    const float* __restrict__ predictions,
    const float* __restrict__ targets,
    float* __restrict__ losses,
    const int batch_size,
    const int input_size
) {
    // Each block handles one sample in the batch
    int sample_idx = blockIdx.x;
    if (sample_idx >= batch_size) return;

    // Shared memory for reductions
    extern __shared__ float sdata[];

    // Pointers to data for this sample
    const float* pred = predictions + sample_idx * input_size;
    const float* targ = targets + sample_idx * input_size;

    // Intermediate sums for dot product and norms
    float thread_dot = 0.0f;
    float thread_pred_norm_sq = 0.0f;
    float thread_targ_norm_sq = 0.0f;

    for (int idx = threadIdx.x; idx < input_size; idx += blockDim.x) {
        float p = pred[idx];
        float t = targ[idx];
        thread_dot += p * t;
        thread_pred_norm_sq += p * p;
        thread_targ_norm_sq += t * t;
    }

    // Reduction for dot product
    sdata[threadIdx.x] = thread_dot;
    __syncthreads();
    for (unsigned int s = blockDim.x / 2; s > 0; s >>= 1) {
        if (threadIdx.x < s) {
            sdata[threadIdx.x] += sdata[threadIdx.x + s];
        }
        __syncthreads();
    }
    float dot_product = sdata[0];

    // Reduction for pred_norm_sq
    sdata[threadIdx.x] = thread_pred_norm_sq;
    __syncthreads();
    for (unsigned int s = blockDim.x / 2; s > 0; s >>= 1) {
        if (threadIdx.x < s) {
            sdata[threadIdx.x] += sdata[threadIdx.x + s];
        }
        __syncthreads();
    }
    float norm_pred = sqrtf(sdata[0] + 1e-8f);

    // Reduction for targ_norm_sq
    sdata[threadIdx.x] = thread_targ_norm_sq;
    __syncthreads();
    for (unsigned int s = blockDim.x / 2; s > 0; s >>= 1) {
        if (threadIdx.x < s) {
            sdata[threadIdx.x] += sdata[threadIdx.x + s];
        }
        __syncthreads();
    }
    float norm_targ = sqrtf(sdata[0] + 1e-8f);

    if (threadIdx.x == 0) {
        float cosine_sim = dot_product / (norm_pred * norm_targ + 1e-8f);
        losses[sample_idx] = 1.0f - cosine_sim;
    }
}
\end{lstlisting}
\end{figure}

\noindent \textbf{\textit{1.9x Speedup} on Level 1 Problem 98 by Deepseek-R1} \\
Torch reference for Cosine Similarity Loss
\begin{figure}[H]
\begin{lstlisting}
self.loss_fn = torch.nn.TripletMarginLoss(margin=margin)
self.loss_fn(anchor, positive, negative)
\end{lstlisting}
\end{figure}
\noindent Another example of a generated kernel using shared memory:
\begin{figure}[H]
\begin{lstlisting}
__global__ void triplet_margin_loss_kernel(
    const float* anchor, 
    const float* positive, 
    const float* negative, 
    float* losses, 
    float margin, 
    int feature_size) 
{
    extern __shared__ float shared_sums[];

    int batch_idx = blockIdx.x;
    int tid = threadIdx.x;

    int offset = batch_idx * feature_size;

    const float* a = anchor + offset;
    const float* p = positive + offset;
    const float* n = negative + offset;

    float a_p_sum = 0.0f;
    float a_n_sum = 0.0f;

    int stride = blockDim.x;
    for (int i = tid; i < feature_size; i += stride) {
        float diff_ap = a[i] - p[i];
        a_p_sum += diff_ap * diff_ap;
        float diff_an = a[i] - n[i];
        a_n_sum += diff_an * diff_an;
    }

    shared_sums[tid] = a_p_sum;
    shared_sums[blockDim.x + tid] = a_n_sum;

    __syncthreads();

    for (int s = blockDim.x / 2; s > 0; s >>= 1) {
        if (tid < s) {
            shared_sums[tid] += shared_sums[tid + s];
            shared_sums[blockDim.x + tid] += shared_sums[blockDim.x + tid + s];
        }
        __syncthreads();
    }

    if (tid == 0) {
        float d_ap = sqrtf(shared_sums[0]);
        float d_an = sqrtf(shared_sums[blockDim.x]);
        losses[batch_idx] = fmaxf(d_ap - d_an + margin, 0.0f);
    }
}
\end{lstlisting}
\end{figure}

\subsection{Iterative Refinement Examples} \label{appendix:iterative-refinement}
\subsubsection{Iteratively Trying new Optimizations}
We provide an example of a kernel that iteratively improves on its existing generation. In the following example, the model attempts new optimizations incorrectly, fixes them, and continue to attempt new optimizations, improving its kernel to faster than the \texttt{torch.compile} baseline ($1.34$ms) but short of the Torch Eager baseline ($0.47$ms).
\\\\
\noindent \textbf{Level 1, Problem 63: 2D convolution with square input and square kernel. DeepSeek-R1 with Execution and Profile Feedback}

\begin{table}[ht]
\centering
\begin{tabular}{lcccccccccc}
\toprule
 \textbf{Turn \#} & \textbf {1} & \textbf{2} & \textbf{3} & \textbf{4} & \textbf{5} & \textbf{6} & \textbf{7} & \textbf{8} & \textbf{9} & \textbf{10} \\
\midrule
\textbf{Compiles?} & $\checkmark$ & \xmark & $\checkmark$ & \xmark & $\checkmark$ & $\checkmark$ & \xmark & $\checkmark$ & \xmark & $\checkmark$ \\
\textbf{Correct?} & $\checkmark$ & \xmark & $\checkmark$ & \xmark & $\checkmark$ & $\checkmark$ & \xmark & $\checkmark$ & \xmark & $\checkmark$ \\
\textbf{Runtime (ms)} & 9.1 & - & 1.57 & - & 1.83 & 1.43 & - & \textbf{1.13} & - & 1.46 \\
\bottomrule
\end{tabular}
\caption{Iterative refinement trajectory of DeepSeek-R1 with execution feedback $E$ and profiler feedback $P$ on Problem 63, Level 1. Torch Eager baseline runs in $0.47$ms and \texttt{torch.compile} runs in $1.34$ms.}
\label{tab:level1-prob63}
\end{table}

\noindent In this example, we see a $8\times$ speedup in average kernel runtime from its initial generation, where the model repeatedly (incorrectly) refines its kernel, fixes the compiler issues using feedback, then continues to attempt more optimizations. The first big jump in performance $(\text{Turn 1} \rightarrow \text{Turn 3})$ occurs because the model decides to launch thread blocks along an output channel dimension, when it originally computed these elements sequentially. The model then attempts to use shared memory in Turn 5, and continues using it, along with texture cache memory with the \texttt{\_\_ldg} instruction in Turns 7 and 8.

\subsubsection{Leveraging Feedback to Correct Kernel Code}
\textbf{Level 2, Problem 73: 2D Convolution with a BatchNorm and a scale factor. DeepSeek-R1 with Execution Feedback}

We provide an example of a kernel that the model struggles to generate correctly, and produces a correct kernel after iterative refinement using execution feedback.

\begin{table}[ht]
\centering
\begin{tabular}{lcccccccccc}
\toprule
 \textbf{Turn \#} & \textbf {1} & \textbf{2} & \textbf{3} & \textbf{4} & \textbf{5} & \textbf{6} & \textbf{7} & \textbf{8} & \textbf{9} & \textbf{10} \\
\midrule
\textbf{Compiles?} & $\checkmark$ & $\checkmark$ & $\checkmark$ & $\checkmark$ & $\checkmark$ & $\checkmark$ & $\checkmark$ & $\checkmark$ & $\checkmark$ & $\checkmark$ \\
\textbf{Correct?} & \xmark & \xmark & \xmark & \xmark & \xmark & \xmark & \xmark & \xmark & \xmark & $\checkmark$ \\
\textbf{Runtime} & - & - & - & - & - & - & - & - & - & \textbf{3.16} \\
\bottomrule
\end{tabular}
\caption{Iterative refinement trajectory of DeepSeek-R1 with execution feedback $E$ on Problem 73, Level 2. Torch Eager baseline runs in $0.105$ms and \texttt{torch.compile} runs in $0.156$ms.}
\label{tab:level2-prob73}
\end{table}

\noindent In the above example, the model continually produces either the wrong output tensor shape or the wrong values and iterates on its kernel using this feedback until the final turn, where it generates a functionally correct, albeit non-performant kernel. We provide another example below that explicitly leverages compiler feedback to fix compiler errors:
\\\\
\noindent \textbf{Level 2, Problem 23: 3D Convolution with a GroupNorm and return the mean across all but the batch dimension. DeepSeek-R1 with Execution Feedback}
\begin{table}[ht]
\centering
\begin{tabular}{lcccccccccc}
\toprule
 \textbf{Turn \#} & \textbf {1} & \textbf{2} & \textbf{3} & \textbf{4} & \textbf{5} & \textbf{6} & \textbf{7} & \textbf{8} & \textbf{9} & \textbf{10} \\
\midrule
\textbf{Compiles?} & $\checkmark$ & $\checkmark$ & $\checkmark$ & $\checkmark$ & \xmark & \xmark & $\checkmark$ & $\checkmark$ & \xmark & $\checkmark$ \\
\textbf{Correct?} & \xmark & \xmark & $\checkmark$ & $\checkmark$ & \xmark & \xmark & $\checkmark$ & $\checkmark$ & \xmark & \xmark \\
\textbf{Runtime} & - & - & 11.4 & 1.36 & - & - & 1.39 & \textbf{1.33} & - & - \\
\bottomrule
\end{tabular}
\caption{Iterative refinement trajectory of DeepSeek-R1 with execution feedback $E$ on Problem 23, Level 2. Torch Eager baseline runs in $1.29$ms and \texttt{torch.compile} runs in $0.719$ms.}
\label{tab:level2-prob23}
\end{table}

In this example, the model attempts to use the \texttt{CUB} library, but incorrectly invokes function calls. The model is then able to correct these errors and write a slightly faster kernel in Turn 8 (see Table~\ref{tab:level2-prob23}).

\subsubsection{Iterative Refinement Never Fixes the Error}
\textbf{Level 1, Problem 54: 3D Convolution square input and square kernel. DeepSeek-R1 with Execution and Profiler Feedback}
\begin{table}[ht]
\centering
\begin{tabular}{lcccccccccc}
\toprule
 \textbf{Turn \#} & \textbf {1} & \textbf{2} & \textbf{3} & \textbf{4} & \textbf{5} & \textbf{6} & \textbf{7} & \textbf{8} & \textbf{9} & \textbf{10} \\
\midrule
\textbf{Compiles?} & $\checkmark$ & $\checkmark$ & $\checkmark$ & $\checkmark$ & $\checkmark$ & $\checkmark$ & $\checkmark$ & $\checkmark$ & $\checkmark$ & $\checkmark$ \\
\textbf{Correct?} & \xmark & \xmark & \xmark & \xmark & \xmark & \xmark & \xmark & \xmark & \xmark & \xmark \\
\textbf{Runtime} & - & - & - & - & - & - & - & - & - & - \\
\bottomrule
\end{tabular}
\caption{Iterative refinement trajectory of DeepSeek-R1 with execution feedback $E$ and profiler feedback $P$ on Problem 54, Level 1. Torch Eager baseline runs in $4.47$ms and \texttt{torch.compile} runs in $4.67$ms.}
\label{tab:level1-prob54}
\end{table}

This problem is particularly interesting because no model is able to consistently produce functional code for this kernel, even with different forms of feedback and profiling information. Interestingly, the example before is an arguably more difficult version of this kernel that fuses the 3D convolution with another operator, and the same model is able to generate functional code for this task. In the example above, the model consistently makes the same mistake and continually generates a functionally incorrect kernel with the same value errors.

\section{Iterative Refinement on Correctness}
\label{appendix:test-time-experiment}
Here we show that \fast{0} across iterative refinement(Section~\ref{sec:iterative-refinement}) configurations at a turn budget of $N=10$ compared to one-shot baseline (Section~\ref{4.1}). We find that models self-correct more effectively with execution feedback $E$, fixing issues especially related to execution errors. Notably, DeepSeek-R1 on Level 1 and 2 can generate a functional kernel on \textgreater 90\% of the tasks given $10$ turns of iterative refinement. However, the remaining incorrect kernels almost always fail due to functional incorrectness, likely because correctness feedback is less granular than execution failure messages.

\begin{table*}[ht]
\centering
\setlength{\tabcolsep}{2pt} % Reduce column spacing
{\small % Only the table content is affected
\begin{tabular}{l|ccc|ccc|ccc}
\toprule
\multirow{3}{*}{\textbf{Method}} & \multicolumn{3}{c|}{\textbf{Level 1}} & \multicolumn{3}{c|}{\textbf{Level 2}} & \multicolumn{3}{c}{\textbf{Level 3}} \\
 & \scriptsize{Llama-3.1} & \scriptsize{DeepSeek} & \scriptsize{Deepseek} & \scriptsize{Llama-3.1} &  \scriptsize{Deepseek} &  \scriptsize{Deepseek} &  \scriptsize{Llama-3.1} &  \scriptsize{Deepseek} &  \scriptsize{Deepseek} \\
 & \small{70B} & \small{V3} & \small{R1} & \small{70B} & \small{V3} & \small{R1} & \small{70B} & \small{V3} & \small{R1} \\
\midrule
Single Attempt (Baseline) & 26\% & 43\% & 67\% & 0\% & 6\% & 62\% & 0\% & 30\% & 8\% \\
\midrule
Iterative Refinement (w G) & 27\% & 48\% & 72\% & 2\% & 7\% & 67\% & 0\% & 36\% & 14\% \\
Iterative Refinement (w G+E) & \textbf{40\%} & \textbf{53\%} & 95\% & \textbf{7\%} & 8\% & 85\% & 18\% & 42\% & \textbf{50\%} \\
Iterative Refinement (w G+E+P) & 36\% & 50\% & \textbf{95\%} & \textbf{7\%} & \textbf{9\%} & \textbf{92\%} & 8\% & \textbf{44\%} & 42\% \\
\bottomrule
\end{tabular}
}
\caption{\textbf{Leveraging execution feedback helps reduce errors:} Here we present the percentage of problems where the LM-generated Kernel is correct for iterative refinement. We note leveraging execution feedback helps the model achieve better correctness \fast{0}, which is the percentage of problems where the model has at least one correct generation up to turn $N=10$. We note the various iterative refinement configurations, leveraging previous Generation $G$, Execution Result $E$, and Timing Profiles $P$.}
\label{table:fast0-method-comparison}
\end{table*}

\section{Few Shot Experiment}
\label{appendix:few-shot-study}

For this experiment, we provide in-context examples of optimization techniques such as fusion, tiling, recompute, and asynchrony to models during kernel generation. As described in Section \ref{subsection:few-shot}, we provide three in-context examples: a fused GELU ~\cite{hendrycks2023gaussianerrorlinearunits}, a tiled matrix multiplication ~\cite{mills2024cuda}, and a minimal Flash-Attention ~\cite{dao2022flashattention, kim2024flashattention} demonstrating effective shared memory I/O management. The prompt used for this experiment is described in Appendix~\ref{appendix:few-shot-study-prompts}. The speedup of these kernels were computed over PyTorch Eager. We compare the performance of these few-shot kernels over the one-shot baseline below.

\begin{table}[h]
\centering

\label{tab:few_shot_model_baseline}
\begin{tabular}{lccccccc}
\toprule
\multicolumn{2}{c}{} & \multicolumn{3}{c}{\textbf{Baseline}} & \multicolumn{3}{c}{\textbf{Few-Shot}}\\ 
    \cmidrule(lr){3-5} \cmidrule(lr){6-8}
\textbf{Model} & \textbf{Level} & \textbf{\fast{1}} & $\textbf{\fast{0}}$ & 
\textbf{Length (chars)} &
$\textbf{\fast{1}}$ & $\textbf{\fast{0}}$ &
\textbf{Length (chars)}\\
\midrule
             & 1 & 3\%  & 27\% & 301018 & 6\%  & 27\% & 360212 \\
Llama 3.1-70B  & 2 & 0\%  & 0\% & 646403 & 0\%  & 0\% & 566668  \\
             & 3 & 0\%  & 0\% & 404596  & 0\%  & 4\% & 485332  \\
\midrule
            & 1 & 10\% & 55\% & 343995 & 6\%  & 39\% & 437768\\
OpenAI o1  & 2 & 24\% & 56\% & 381474 & 16\% & 39\% & 432800 \\
            & 3 & 12\% & 56\% & 260273 & 8\%  & 22\% & 364551 \\
\bottomrule
\end{tabular}

\caption{Comparison of the Section \ref{4.1} baseline and few-shot prompting performance across models. We examine the $\textbf{\fast{0}}$, $\textbf{\fast{1}}$, and cumulative character length of generated kernels per level.}

\end{table}
\FloatBarrier

\noindent 77\% of matrix multiplication problems in Level 1 achieves a speedup over the one-shot baseline through tiling. The runtime comparison for each GEMM variant is presented below.
\begin{table}[H]
\centering
\label{tab:few_shot_kernel_fusion_results_level1}
\begin{tabular}{lccc}
    \toprule
    \textbf{Problem Name} & \textbf{Baseline (ms)} & \textbf{Few-Shot (ms)} & \textbf{Ref Torch (ms)} \\
    \midrule
    3D Tensor Matrix Multiplication & 20.9 & 7.71 & 1.45 \\
    Matmul for Upper-Triangular Matrices & 14 & 5.39 & 2.98 \\
    Matrix Scalar Multiplication & 1.19 & 0.811 & 0.822 \\
    Standard Matrix Multiplication & 3.39 & 2.46 & 0.397 \\
    Matmul with Transposed Both & 3.44 & 2.67 & 0.412 \\
    Matmul with Transposed A & 3.61 & 2.99 & 0.384 \\
    4D Tensor Matrix Multiplication & 366 & 338 & 36 \\
    Tall Skinny Matrix Multiplication & 3.39 & 3.59 & 1.9 \\
    Matmul with Diagonal Matrices & 0.221 & 0.237 & 2.83 \\
    \bottomrule
\end{tabular}

\caption{Performance comparison of the Section \ref{4.1} baseline and few-shot prompting in level 1 matrix multiplication problems.}

\end{table}

\noindent Few-shot kernels generated for the following problems in level 2 outperformed PyTorch Eager through aggressive shared memory I/O management.
\begin{table}[H]
\centering
\label{tab:few_shot_kernel_fusion_results_level2}
\begin{tabular}{lrrr}
\toprule
\textbf{Problem Name} & \textbf{Baseline (ms)} & \textbf{Few-Shot (ms)} & \textbf{Ref Torch (ms)} \\
\midrule
Conv2d InstanceNorm Divide & 0.514 & 0.0823 & 0.0898 \\
Gemm GroupNorm Swish Multiply Swish & 0.124 & 0.0542 & 0.0891 \\
Matmul Min Subtract & 0.0651 & 0.0342 & 0.0397 \\
Matmul GroupNorm LeakyReLU Sum & 0.0935 & 0.0504 & 0.072 \\
ConvTranspose3d Swish GroupNorm HardSwish & 33.3 & 29.6 & 35.2 \\
ConvTranspose2d Mish Add Hardtanh Scaling & 0.235 & 0.209 & 0.243 \\
ConvTranspose3d Add HardSwish & 15.6 & 14.1 & 22.2 \\
ConvTranspose2d Add Min GELU Multiply & 0.365 & 0.349 & 0.4 \\
ConvTranspose2d BiasAdd Clamp Scaling Clamp... & 0.3 & 0.31 & 0.368 \\
Conv2d GroupNorm Tanh HardSwish ResidualAdd... & 0.124 & 0.129	& 0.154 \\
Conv2d ReLU HardSwish & 0.0681 & 0.0711 & 0.0768 \\
\bottomrule
\end{tabular}
\caption{Performance comparison of the Section \ref{4.1} baseline and few-shot prompting in level 2 for problems whose few-shot kernels outperform PyTorch Eager.}
\end{table}

\section{Cross-Hardware Case Study}
\label{appendix:cross-hardware-study}
\subsection{Evaluation across different hardware}

To evaluate how generated kernels fare across different hardware platforms, we utilize a number of different NVIDIA GPUs that span different micro-architectures and capabilities. The specific details for each is provided in Table \ref{tab:gpu_specifications}.

\begin{table}[ht]
\centering
\begin{tabular}{lllllll}
\hline

\textbf{Provider} & \textbf{GPU Type} & \textbf{Memory} & \textbf{Power} & \textbf{Microarchitecture} & \textbf{FP16 TFLOPS} & \textbf{Memory Bandwidth}\\ 
\hline

Baremetal& NVIDIA L40S       & 48 GB           & 300W           & Ada      & 362.05 & 864 GB/s                 \\ 
Baremetal& NVIDIA H100       & 80 GB           & 700W           & Hopper      & 989.5 & 3350 GB/s              \\ 
Serverless& NVIDIA L40S       & 48 GB           & 350W           & Ada       & 362.05 & 864 GB/s                \\ 
Serverless& NVIDIA A100       & 42 GB           & 400W           & Ampere    & 312 & 1935 GB/s                \\ 
Serverless& NVIDIA L4         & 24 GB           & 72W            & Ada       & 121 & 300 GB/s                \\ 
Serverless& NVIDIA T4         & 16 GB           & 70W            & Turing     & 65 & 300 GB/s               \\ 
Serverless& NVIDIA A10G       & 24 GB           & 300W           & Ampere      & 125 & 600 GB/s              \\ 
\hline

\end{tabular}
\caption{Specifications of different GPUs, including memory, power consumption, micro-architecture, FP16 TFLOPS, memory bandwidth, and their providers.}
\label{tab:gpu_specifications}
\end{table}

\noindent We ran the same set of kernels generated in Section \ref{4.1} on a variety of hardware (as listed in Table \ref{tab:gpu_specifications}). We computed the \fast{1} speedup against the PyTorch Eager baseline profiled on that particular hardware platform in Table \ref{tab:speedup-hardware-comparison}.

\begin{table}[H]
\centering
\begin{tabular}{llccc}
\toprule
\textbf{Level} & \textbf{GPUs} & \textbf{Llama-3.1-70b-Inst} & \textbf{DeepSeek-V3} & \textbf{DeepSeek-R1} \\
\midrule
\multirow{6}{*}{
\textbf{1}}
 & \textbf{L40S} & 3\% & 6\% & 12\% \\
 & \textbf{H100} & 2\% & 7\% & 16\% \\
 & \textbf{A100} & 3\% & 7\% & 16\%\\
 & \textbf{L4} & 2\% & 4\% & 15\% \\
 & \textbf{T4} & 3\% & 7\% & 22\% \\
 & \textbf{A10G} & 2\% & 7\% & 12\% \\
\midrule
\multirow{6}{*}{\textbf{2}}
 & \textbf{L40S} & 0\% & 4\% & 36\% \\
 & \textbf{H100} & 0\% & 4\% & 42\%\\
 & \textbf{A100} & 0\% & 4\% & 38\%\\
 & \textbf{L4} & 0\% & 4\% & 36\% \\
 & \textbf{T4} & 0\% & 4\% & 46\% \\
 & \textbf{A10G} & 0\% & 4\% & 47\% \\
\midrule
\multirow{6}{*}{\textbf{3}}
 & \textbf{L40S} & 0\% & 8\% & 2\% \\
 & \textbf{H100} & 0\% & 10\% & 2\% \\
 & \textbf{A100} & 0\% & 8\% & 2\% \\
 & \textbf{L4} & 0\% & 6\% & 2\% \\
 & \textbf{T4} & 0\% & 10\% & 2\% \\
 & \textbf{A10G} & 0\% & 10\% & 0\% \\
\bottomrule
\end{tabular}
\caption{KernelBench result across multiple hardware types: Speedup (\fast{1}) over Torch Eager comparison of GPUs across different models and levels. The kernels used across different GPUs are the same as the ones generated for Single Attempt \textbf{without} hardware/platform specific information.}
\label{tab:speedup-hardware-comparison}
\end{table}

\noindent Based on the increased variability in \fast{1} score for DeepSeek R1 as described in Section \ref{section:perf-across-hardware} and Table \ref{tab:speedup-hardware-comparison}, we plot the individual speedups for each problem (in Levels 1 and 2) across different GPUs. Speedup is computed against PyTorch Eager and there is a horizontal line at $y = 1.0$ to mark the cutoff for \fast{1}.
\begin{figure}[H]
    \centering
    \includegraphics[scale=0.45]{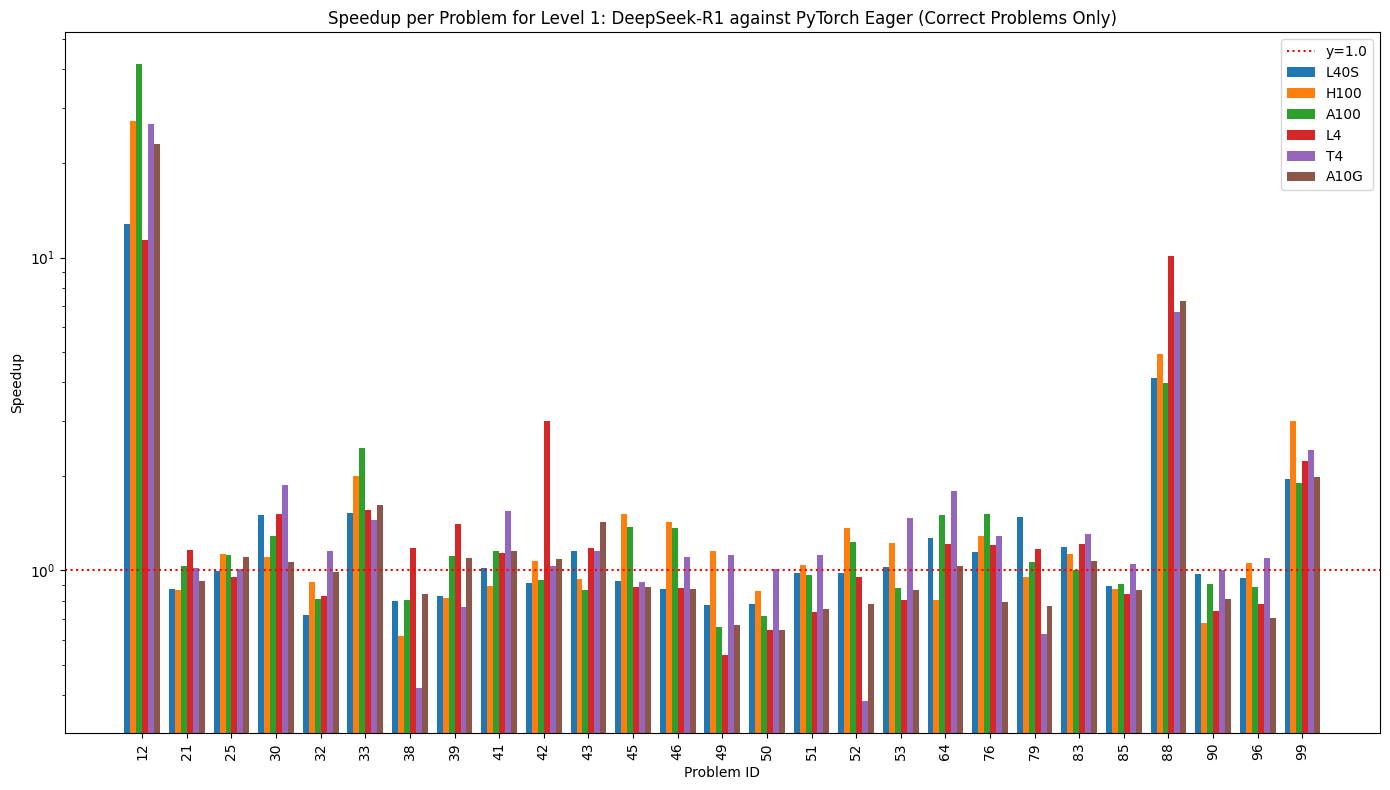}
    \caption{Speedup comparison across different GPUs for DeepSeek R1 on Level 1 (log scale).}
    \label{fig:r1_eager_level1_hw}
\end{figure}

\begin{figure}[H]
    \centering
    \includegraphics[width=\textwidth]{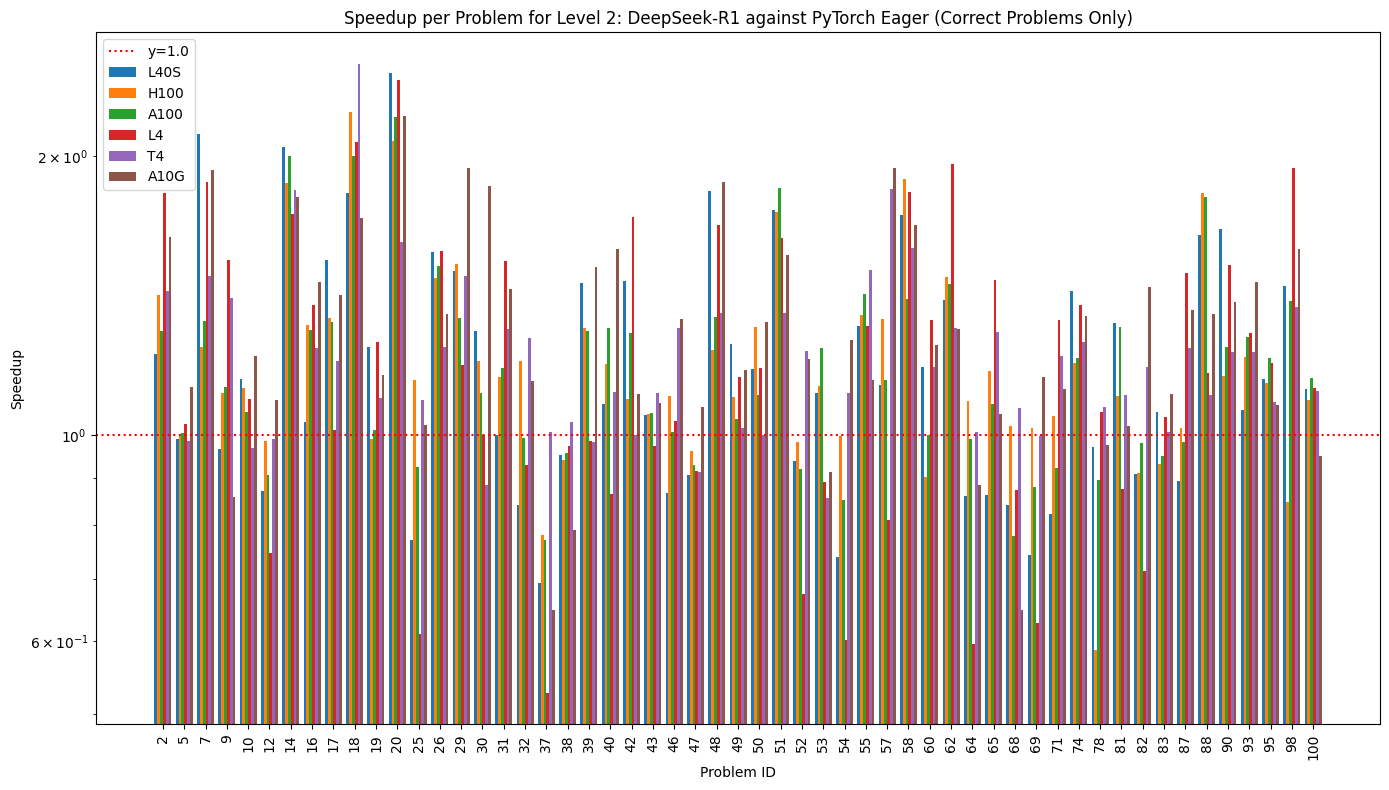}
    \caption{Speedup comparison across different GPUs for DeepSeek-R1 on Level 2 (log scale).}
    \label{fig:r1_eager_level2_hw}
\end{figure}

\subsection{Effect of Providing Hardware Information}

We provided hardware-specific information such as the GPU's micro-architecture, TFLOPS, and memory bandwidth (as shown in Table \ref{tab:gpu_specifications}) to models during kernel generation. These generations were then profiled on their respective hardware platforms. The speedup in terms of \fast{1} over PyTorch Eager is provided below in Table \ref{tab:speedup-hardware-comparison-prompting}.

\begin{table}[H]
\centering
\begin{tabular}{llcccl}
\toprule
\textbf{Level} & \textbf{GPUs} & \textbf{Llama-3.1-70b-Inst} & \textbf{DeepSeek-V3} & \textbf{DeepSeek-R1}  & \textbf{OpenAI-o1}\\
\midrule
\multirow{3}{*}{
\textbf{1}}
& \textbf{L40S} & 2\% & 5\% & 14\%  &13\%\\
& \textbf{H100} & 3\% & 6\% & 14\%  &9\%\\
& \textbf{A100} & 4\% & 4\% & 16\%  &9\%\\
\midrule
\multirow{3}{*}{
\textbf{2}}
& \textbf{L40S} & 1\% & 4\% & 33\%  &25\%\\
& \textbf{H100} & 2\% & 2\% & 35\%  &33\%\\
& \textbf{A100} & 2\% & 1\% & 31\%  &32\%\\
\midrule
\multirow{3}{*}{
\textbf{3}}
& \textbf{L40S} & 0\% & 2\% & 0\%  &10\%\\
& \textbf{H100} & 2\% & 8\% & 0\%  &12\%\\
& \textbf{A100} & 0\% & 0\% & 2\%  &8\%\\
\bottomrule
\end{tabular}

\caption{Providing hardware specifications in context: Speedup (\fast{1}) over Torch Eager comparison of GPUs across different models and levels.}
\label{tab:speedup-hardware-comparison-prompting}
\end{table}

As described in Section \ref{subsection:hardware-prompting-study}, DeepSeek-R1 begins to generate kernels with  warp matrix multiply-accumulate (\texttt{wmma}) instructions. We show the generated kernel for Problem 1 Level 1 when provided with H100 specifications in context below.

\begin{figure}
\begin{lstlisting}
#include <torch/extension.h>
#include <cuda_fp16.h>
#include <cuda_runtime.h>
#include <mma.h>

using namespace nvcuda;

__global__ void hgemm_kernel(const half* A, const half* B, half* C, int N) {
    const int WARPS_PER_BLOCK = 8;
    const int WMMA_M = 16;
    const int WMMA_N = 16;
    const int WMMA_K = 16;
    
    __shared__ half Ashared[WARPS_PER_BLOCK][WMMA_M][WMMA_K];
    __shared__ half Bshared[WARPS_PER_BLOCK][WMMA_K][WMMA_N];

    int warpId = threadIdx.x / 32;
    int laneId = threadIdx.x % 32;
    
    int tileM = (blockIdx.y * WARPS_PER_BLOCK + warpId) * WMMA_M;
    int tileN = blockIdx.x * WMMA_N;
    
    wmma::fragment<wmma::matrix_a, WMMA_M, WMMA_N, WMMA_K, half, wmma::row_major> a_frag;
    wmma::fragment<wmma::matrix_b, WMMA_M, WMMA_N, WMMA_K, half, wmma::row_major> b_frag;
    wmma::fragment<wmma::accumulator, WMMA_M, WMMA_N, WMMA_K, half> c_frag;
    wmma::fill_fragment(c_frag, __float2half(0.0f));

    for (int k = 0; k < N; k += WMMA_K) {
        int loadAidx = laneId + (tileM + (laneId / WMMA_K)) * N + k + (laneId % WMMA_K);
        int loadBidx = laneId + (k + (laneId / WMMA_N)) * N + tileN + (laneId % WMMA_N);
        
        if (tileM + (laneId / WMMA_K) < N && k + (laneId % WMMA_K) < N)
            Ashared[warpId][laneId / WMMA_K][laneId % WMMA_K] = A[loadAidx];
        else
            Ashared[warpId][laneId / WMMA_K][laneId % WMMA_K] = __float2half(0.0f);

        if (k + (laneId / WMMA_N) < N && tileN + (laneId % WMMA_N) < N)
            Bshared[warpId][laneId / WMMA_N][laneId % WMMA_N] = B[loadBidx];
        else
            Bshared[warpId][laneId / WMMA_N][laneId % WMMA_N] = __float2half(0.0f);

        __syncthreads();

        wmma::load_matrix_sync(a_frag, &Ashared[warpId][0][0], WMMA_K);
        wmma::load_matrix_sync(b_frag, &Bshared[warpId][0][0], WMMA_N);
        wmma::mma_sync(c_frag, a_frag, b_frag, c_frag);
        
        __syncthreads();
    }

    int storeCidx = (tileM + (laneId / WMMA_N)) * N + tileN + (laneId % WMMA_N);
    if (tileM + (laneId / WMMA_N) < N && tileN + (laneId % WMMA_N) < N)
        C[storeCidx] = c_frag.x[laneId];
}

torch::Tensor hgemm_cuda(torch::Tensor A, torch::Tensor B, int N) {
    auto C = torch::zeros({N, N}, A.options().dtype(torch::kFloat16));
    
    const int WARPS_PER_BLOCK = 8;
    dim3 grid((N + 15) / 16, (N + 15) / (16 * WARPS_PER_BLOCK));
    dim3 block(32 * WARPS_PER_BLOCK);
    
    hgemm_kernel<<<grid, block>>>(A.data_ptr<half>(), B.data_ptr<half>(), C.data_ptr<half>(), N);
    return C;
}
\end{lstlisting}
\caption{A CUDA kernel generated by DeepSeek-R1 for Level 1 Problem 1 when provided with hardware-specific information on the H100 GPU.}
\label{fig:example_r1_generated_kernel_hw}
\end{figure}
\FloatBarrier

\section{High-Throughput Evaluation System}
\label{appendix/eval-system}
\subsection{Single-shot Experiments: Batched Kernel Generation}
\begin{figure}[h]
    \centering
    \includegraphics[width=0.75\linewidth]{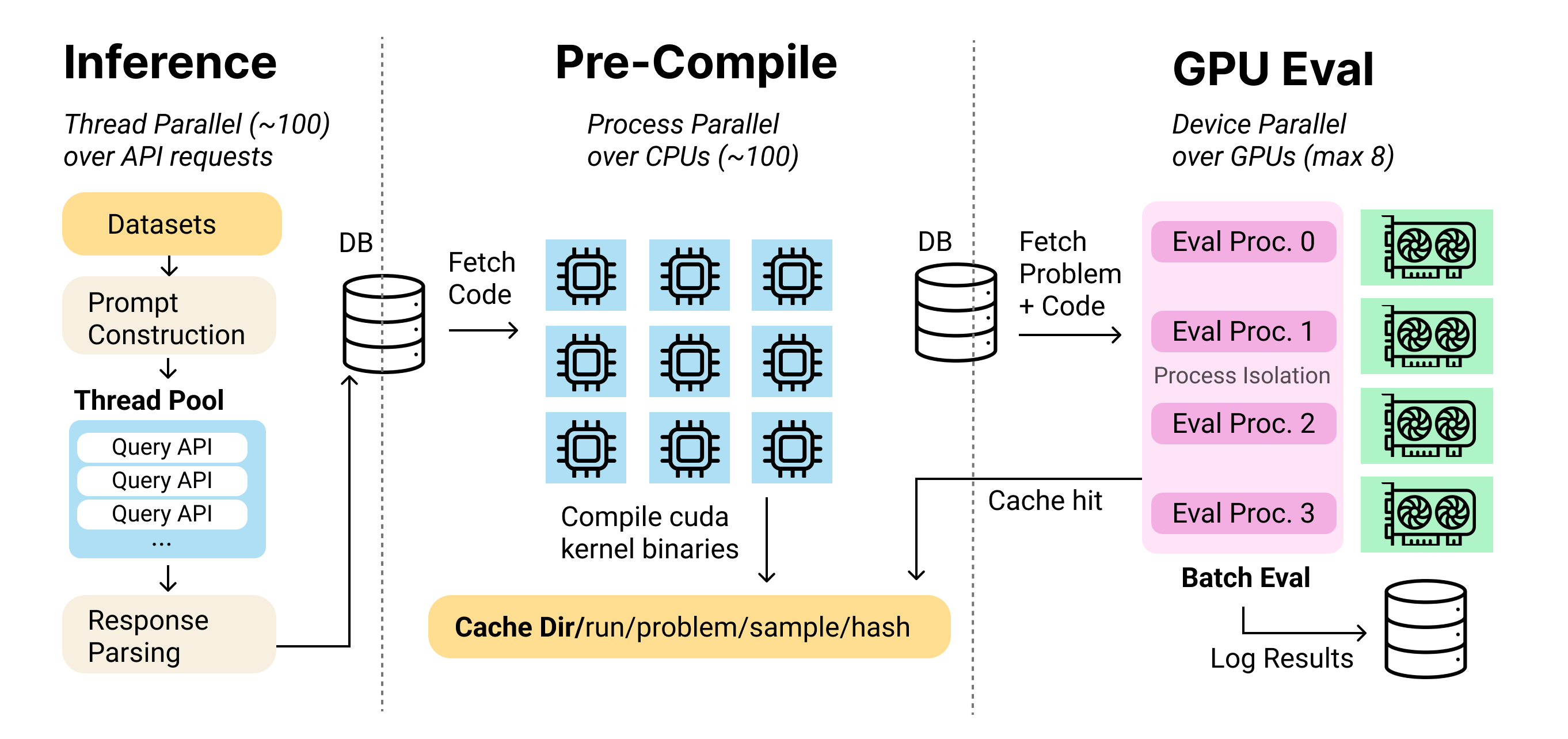}
    \caption{\textbf{KernelBench provide a high throughput kernel generation and evaluation system}. We parallelized generation, compilation, and evaluation of kernels across CPUs and GPUs.}
    \label{fig:parallel-eval-system}
\end{figure}

Given the high volume of GPU kernels to evaluate, we build a fast and highly-parallelized evaluation system, where we separate into the kernel generation and evaluation process into 3 stages, as shown in Figure~\ref{fig:parallel-eval-system}.
\begin{itemize}
    \item \textbf{Inference:} We query LMs in parallel and store the generated kernel.
    \item \textbf{CPU Pre-Compile:} We compile the model-generated kernels with \texttt{nvcc} for a specified hardware into a binary, parallelized on CPUs and each kernel binary is saved to their individual specific directory for caching. 
    \item \textbf{GPU Evaluation:} With the kernel binary already built on CPU, we focus on evaluating multiple kernels in parallel across multiple GPU devices. However, to ensure accurate kernel timing, we only evaluate one kernel at time on one device. 
\end{itemize}

\subsection{Iterative Refinement Experiments: GPU Orchestrator System}

Based on the single-shot system, we also design a platform to handle multiple iterative refinement experiments at once. We treat each iterative refinement experiment as a finite state machine, where the states are LM-based kernel generation, pre-compilation, kernel execution, and profiling. The transitions are based on environment feedback, and can change based on different experiment setups.

Our system was run on a node with $8$ available GPUs. Unlike the single-shot system, batching each generation and kernel execution is highly inefficient -- thus, we design a pipelined, multiprocessing system with a GPU orchestrator with the following characteristics:

\begin{itemize}
    \item \textbf{CPU Parallelism:} The orchestrator spawns multiple independent processes that each handle an independent task in KernelBench. These processes run the multi-turn state machine logic for the iterative refinement experiments -- only the kernel execution state requires acquiring a GPU.
    \item \textbf{Acquiring GPUs:} The GPU orchestrator keeps a separate process running that handles which processes can acquire a GPU using semaphores. Processes can request a GPU from this process when it is ready to execute and evaluate kernel code. We try to minimize process control over a GPU to maximize resource throughput, given a system with a limited number of available GPUs.
    \item \textbf{Pre-compiling on the CPU:} To avoid processes hogging GPU time, we pre-compile kernels with \texttt{nvcc} on the CPU for a specified hardware into a binary. We also did this same trick for the single-shot system, but for separate reasons.
    \item \textbf{Evaluating Kernels on the GPU:} The only state where the finite state machine uses the GPU is for kernel execution and profiling. We found that waiting on GPUs is the primary bottleneck in the orchestrator, so we designed the orchestrator to maximize device occupancy.
\end{itemize}

\noindent The system generally supports overlapping the generation of kernel code and the execution of already-generated kernel code. There are also several unavoidable errors such as CUDA illegal memory accesses and deadlocks due to faulty kernel generations that the orchestrator solves by releasing and spawning new processes when encountered, and we wrote specifically handlers to ensure these errors are properly captured without crashing the orchestrator itself.

\subsection{UI: Visualizing Kernel Generation Trajectories}

To qualitatively observe the generated and compare them across techniques, we design an interface to easily visualize them. We provide this as part of the KernelBench framework.

\begin{figure}[h]
    \centering
    \includegraphics[width=0.9\linewidth]{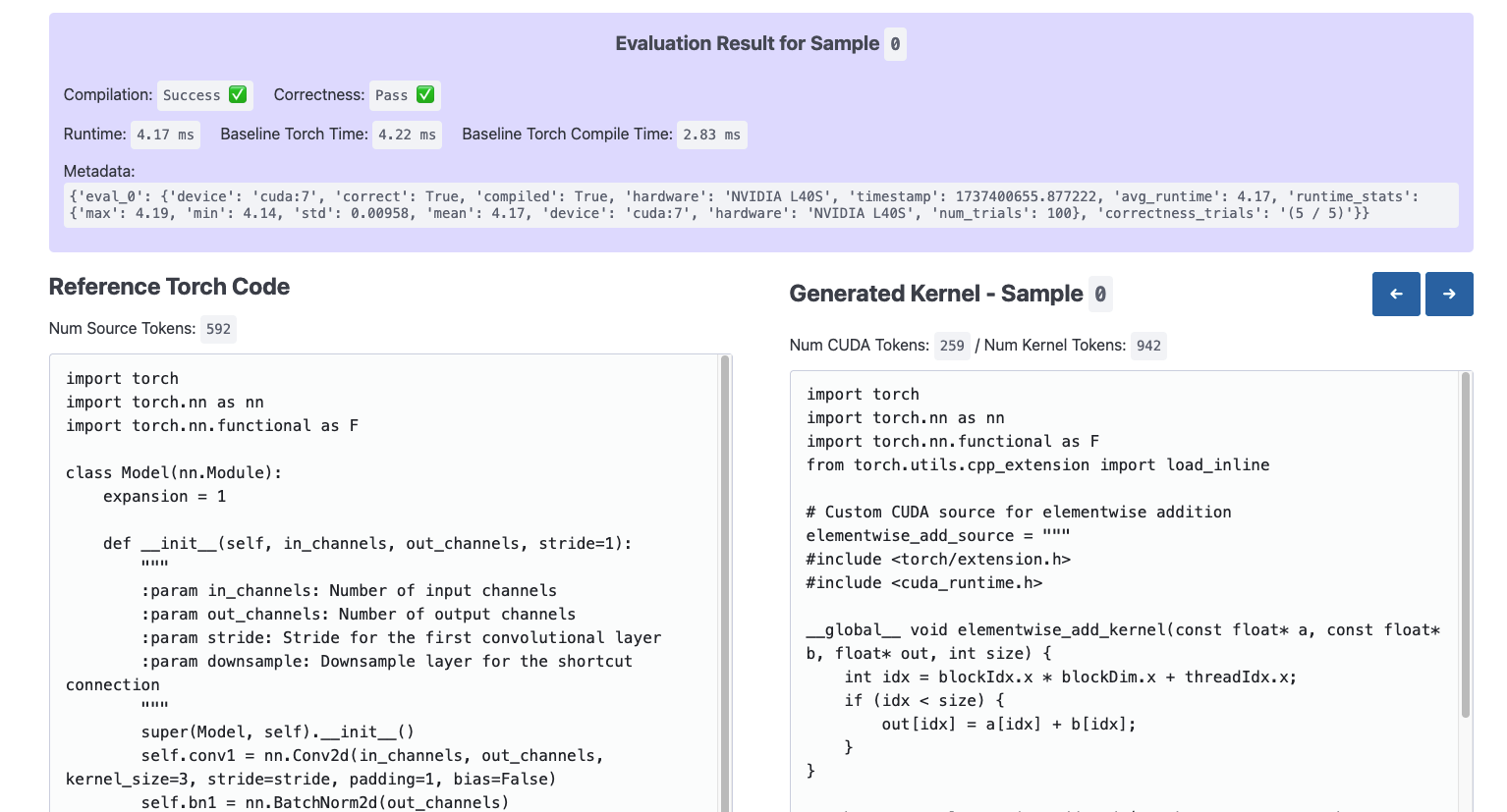}
    \caption{\textbf{We provide a visual interface for kernel inspection}. This allows us to easily examine kernel content, its performance, and compare across various techniques and configurations.}
    \label{fig:kernel_inspect}
\end{figure}

% The $\mathtt{\backslash onecolumn}$ command above can be kept in place if you prefer a one-column appendix, or can be removed if you prefer a two-column appendix.  Apart from this possible change, the style (font size, spacing, margins, page numbering, etc.) should be kept the same as the main body.
%%%%%%%%%%%%%%%%%%%%%%%%%%%%%%%%%%%%%%%%%%%%%%%%%%%%%%%%%%%%%%%%%%%%%%%%%%%%%%%
%%%%%%%%%%%%%%%%%%%%%%%%%%%%%%%%%%%%%%%%%%%%%%%%%%%%%%%%%%%%%%%%%%%%%%%%%%%%%%%

\end{document}